\documentclass[runningheads]{llncs}
\usepackage{graphicx}

\usepackage{tikz}
\usepackage{comment}
\usepackage{amsmath,amssymb} 
\usepackage{color}
\usepackage{caption}
\usepackage[accsupp]{axessibility}  

\usepackage{multirow}
\usepackage[nodisplayskipstretch]{setspace}
\usepackage{array, multirow}
\usepackage{amsmath,amssymb,stmaryrd}   
\usepackage{multicol}

\newcommand\blfootnote[1]{%
  \begingroup
  \renewcommand\thefootnote{}\footnote{#1}%
  \addtocounter{footnote}{-1}%
  \endgroup
}
\begin{document}
\pagestyle{headings}
\mainmatter
\def\ECCVSubNumber{20}  

\title{Hybrid Transformer Based Feature Fusion for Self-Supervised Monocular Depth Estimation} 

\titlerunning{ECCV-22 submission ID \ECCVSubNumber} 
\authorrunning{ECCV-22 submission ID \ECCVSubNumber} 
\author{Anonymous ECCV submission}
\institute{Paper ID \ECCVSubNumber}

\titlerunning{Hybrid Transformer Based Depth Estimation}
%
\author{Snehal Singh Tomar*  \and
Maitreya Suin* \and
A. N. Rajagopalan \\
\email{snehal@smail.iitm.ac.in, maitreyasuin21@gmail.com, raju@ee.iitm.ac.in}
}
\authorrunning{Tomar et al.}
%

\institute{Indian Institute Of Technology Madras, India}
\maketitle
\blfootnote{*Equal contribution.}
\begin{abstract}
With an unprecedented increase in the number of agents and systems that aim to navigate the real world using visual cues and the rising impetus for 3D Vision Models, the importance of depth estimation is hard to understate. While supervised methods remain the gold standard in the domain, the copious amount of paired stereo data required to train such models makes them impractical. Most State of the Art (SOTA) works in the self-supervised and unsupervised domain employ a ResNet-based encoder architecture to predict disparity maps from a given input image which are eventually used alongside a camera pose estimator to predict depth without direct supervision. The fully convolutional nature of ResNets makes them susceptible to capturing per-pixel local information only, which is suboptimal for depth prediction. Our key insight for doing away with this bottleneck is to use Vision Transformers, which employ self-attention to capture the global contextual information present in an input image. Our model fuses per-pixel local information learned using two fully convolutional depth encoders with global contextual information learned by a transformer encoder at different scales. It does so using a mask-guided multi-stream convolution in the feature space to achieve state-of-the-art performance on most standard benchmarks.
\end{abstract}

\section{Introduction}
The ability of humans to perceive 3D geometry in the complex environment around them is an indispensable sensory endowment. All agents seeking human-like performance in real-world navigation must learn to emulate this capability. Depth estimation is a crucial component of understanding 3D geometry. Traditional Computer Vision techniques estimate depth by calculating disparity over stereo image pairs with known correspondences using prior knowledge of intrinsic camera parameters. Most modern-day autonomous agents synergize image data with information from other sensor modalities such as LiDARs, RGB-D cameras, short-range RADARs, and Ultrasonic Sensors for accurate depth estimation. Acquiring such data is expensive in terms of capital and human effort. Monocular Depth Estimation is the task of predicting the depth information of a scene using a single RGB image as input. Although supervised methods (\cite{lidepthtoolbox2022}, \cite{Bhat_2021_CVPR}, \cite{Kopf_2021_CVPR}) have presented the best results for the task, getting ground truth depth data to make them work is challenging. SOTA self-supervised and unsupervised methods (\cite{monodepth2}, \cite{klingner2020selfsupervised}, \cite{Johnston_2020_CVPR}) that perform Monocular Depth Estimation are inspired from \cite{Zhou_2017_CVPR} and make use of fully convolutional architectures for predicting disparity from a given image. They also have a separate network for predicting camera pose, the predictions of which are clubbed with the depth predictions and prior knowledge of intrinsic camera parameters to warp a given frame in a monocular video sequence onto another temporally consistent frame. These temporally consistent frames thus, serve as a surrogate for direct depth map supervision to perform this ill-posed task. 

\section{Related Works}
\textbf{Self-Supervised Monocular Depth Estimation:} The advent of deep learning architectures in Computer Vision made Monocular Depth Estimation a reality. Monocular Depth Estimation had been attempted by several works (\cite{NIPS2014_7bccfde7}, \cite{Fu_2018_CVPR}) primarily in a supervised fashion only until \cite{Zhou_2017_CVPR} presented a reliable Structure from Motion (SfM) inspired self-supervised framework. A category of literature (\cite{monodepth2}, \cite{Johnston_2020_CVPR}, \cite{klingner2020selfsupervised}, \cite{Guizilini_2020_CVPR}) has built upon the foundation laid by \cite{Zhou_2017_CVPR} by making important improvements to the Depth and Pose estimation networks. Amongst these, \cite{monodepth2} is a seminal work that makes several important contributions, such as a robust architecture based on ResNet18 \cite{He_2016_CVPR}, multi-scale depth estimation, and per-pixel minimum reprojection loss.                
\newline\textbf{Representation Learning for Augmenting Depth Estimation:}
Both supervised and self-supervised Monocular Depth Estimation methods have recently started to look beyond ground-truth depth maps and monocular video, respectively, as a supervision signal for their models without resorting to multimodal sensor fusion. They learn rich data representations for related downstream tasks such as semantic segmentation or classification and fuse them with the representations learned for depth estimation. Representations learnt for semantic segmentation has been used extensively by such models (\cite{9156944}, \cite{8954386}, \cite{Kumar_2021_WACV}, \cite{Saeedan_2021_WACV}). Features learned for Monocular Depth Estimation usually aid Semantic Segmentation models (\cite{9157827}, \cite{HE2021251}). The methods described in this section differ from methods that use additional supervision in the form of stereo images or data from additional sensor modalities.              
\newline\textbf{Feature Fusion in Latent Space}
Much research has gone into finding optimal fusion techniques for combining latent representations learned via different architectures, latent representations learned for different tasks, and latent representations learned over different data modalities. \cite{9382861} proposes \textit{Feature Attention Fusion} and \textit{Multi-Scale Feature Fusion Dense Pyramid} for Monocular Depth Estimation. \cite{DBLP:journals/corr/abs-2012-10296} proposes combining representations learnt via both 2D and 3D convolutions. \cite{9578558} explores latent space feature fusion for depth map aggregation. \cite{Yang_2021_ICCV} fuses latent space features corresponding to global and local attributes for image retrieval. The success of such approaches has motivated our method, which learns from both global and local receptive fields for the current task. We describe our fusion module in section 3.1.
\newline\textbf{Vision Transformers}
Transformer Models, originally proposed for Natural Language tasks \cite{vaswani2017attention} have revolutionized the way sequence-to-sequence problems are attempted. Vision Transformers (ViTs) (\cite{touvron2021training}, \cite{dosovitskiy2020image} , \cite{dosovitskiy2021an}) have adopted the idea by splitting images into patches and learning their representations as a sequence. A transformer architecture can extract information from an entire image in a single step. The recent past has seen a steep surge in the utility of ViTs for standard vision tasks such as  image recognition \cite{dosovitskiy2021an}, segmentation (\cite{Xu_2022_CVPR}, \cite{Xu_2022_CVPR_1}, \cite{Gu_2022_CVPR}, \cite{Botach_2022_CVPR}), object detection \cite{He_2022_CVPR}, restoration \cite{wang2022uformer}, and inpainting \cite{Dong_2022_CVPR} . Many recent methods have applied self-attention within local regions \cite{liang2021swinir,wang2022uformer} to limit the computational requirement. In our work, we adopt the transposed attention-based transformer design of \cite{zamir2022restormer} for its efficient operation. 

Although convolutional networks excel at processing local details, they lack the required adaptability and contextual information. Correlation among different regions should result in better depth estimation accuracy for objects of different shapes and scales. Recently, \cite{Johnston_2020_CVPR} embedded a self-attention layer within a CNN. But, it limits the potential as it was applied only on the CNN-generated feature maps of the lowest resolution. Although \cite{ranftl2021vision} utilized a pure transformer architecture for depth prediction, it was trained using GT supervision. The behavior of transformers and their relation with CNN is rather unexplored in the self-supervised domain. A pure transformer architecture potentially lacks the required finesse in processing the local details, such as object boundaries. Thus, we design a hybrid transformer architecture that exploits the long-range modeling capability of the transformer while maintaining the local pixel-level accuracy using parallel convolutional branches operating at different resolutions. A detailed description of the architecture is given in the following section. 
\section{Method}
Our goal is to generate the per-pixel depth map of a single frame at inference time. To learn such a function, usually, image-to-image CNNs are trained in a supervised manner where the supervision of the ground-truth depth map is available while training. But, practically, creating such annotated datasets for a large number of scenes is costly. Instead, we follow the self-supervised framework of \cite{monodepth2}, which only requires short image sequences of a scene captured by a moving camera. In such an approach, the primary supervision comes from the task of novel view synthesis. A depth-prediction network is trained to predict the depth of a scene. A separate pose-estimation is required while training that produces the relative camera pose given two consecutive frames. Given the depth map and the relative pose, the appearance of the target frame seen from a different camera pose is synthesized. As the networks are trained to perform view synthesis using the predicted depth map of the current scene, it inherently learns to extract the underlying scene structure. Thus, the whole framework can be trained in a self-supervised manner without requiring GT annotations. Note that the pose estimation is only needed for training to perform view-synthesis. The depth estimation network alone is sufficient for predicting the depth at inference.

\begin{figure}[t]
    \centering
    \includegraphics[width = 0.95\textwidth]{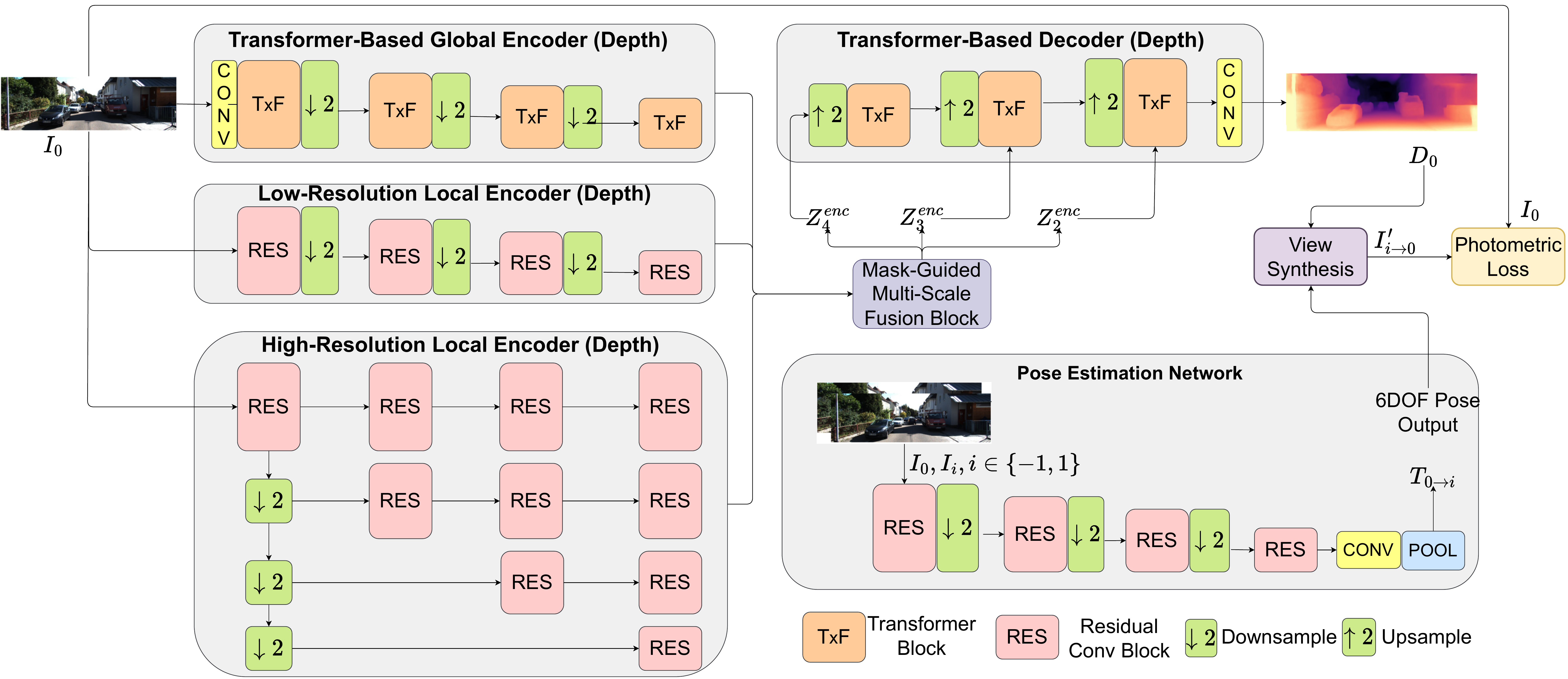}
    \caption{An overview of our hybrid-transformer based self-supervised depth prediction framework. Three parallel encoder branches are shown on the left. The information is fused using the mask-guided multi-scale fusion block, and finally the depth decoder estimates the depth map. The pose network is derived from ResNet18 \cite{He_2016_CVPR} architecture.}
    \label{fig:arch}
    \vspace{-2mm}
\end{figure}

Our overall approach is shown in Fig. \ref{fig:arch}. For the depth-estimation encoder, we first deploy a global attention-based branch focusing on long-range information modeling. Next, to capture the local details better, we deploy two local high-resolution (HRL) and low-resolution (LRL) branches, focusing on either overall scene content (LRL) or sharper pixel details (HRL). Encoder features from these three encoder branches are fused using a multi-scale fusion block, which utilizes a spatial masking operation to extract the most useful information. Next, a transformer-based decoder module uses the fused encoded information to predict the depth map of the scene. We use a simpler ResNet-based pose network as we found it adequate for the relatively easier pose estimation task. In the following section, we first describe the structure of the depth and pose estimation networks, followed by a discussion of the training strategies.

\subsection{Depth Estimation Network}
The depth estimation network has an encoder-decoder framework. We deploy three parallel encoders for capturing local as well as global information. First, we will discuss the individual encoder branches, followed by the description of the information fusion module and depth prediction in the decoder. 
\subsubsection{Transformer-Based Global Encoder:}
Let $I \in \mathbb{R}^{3 \times H \times W}$ be the input frame. We first apply a single convolutional layer in the global branch to generate the initial shallow feature embeddings $X \in \mathbb{R}^{C \times H \times W}$. Next, $X$ is passed through four encoder levels with an increasing number of feature maps and decreasing spatial resolution. For downsampling, we apply pixel-unshuffle operation \cite{shi2016real}. Each level of the encoder contains multiple Vision-Transformer blocks. The operation of one such block is shown in Fig. \ref{fig:txfusion} (a). Given the input tensor $\hat{X} \in \mathbb{R}^{\hat{C} \times \hat{H} \times \hat{W}}$ at any particular level, we first generate three different embeddings: query ($Q \in \mathbb{R}^{d_q \times \hat{H} \times \hat{W}}$), key ($K \in \mathbb{R}^{d_q \times \hat{H} \times \hat{W}}$), and value ($V \in \mathbb{R}^{d_v \times \hat{H} \times \hat{W}}$). We apply standard convolutional operations on $X$ to generate $Q, K, V$. In the original transformer architecture \cite{vaswani2017attention}, the correlations between all possible elements were calculated using $Q^TK$, which will result in a huge $\hat{H}\hat{W} \times \hat{H}\hat{W}$ matrix for images. Instead, we follow the design of \cite{zamir2022restormer} and generate a transposed-attention map $QK^T$ significantly reducing the memory requirement. Next, each pixel value of the output feature map $Y$ is updated using global information as   
\begin{equation}
\label{eq:att}
Y = \hat{X} + \text{conv}_{1\times1}(V \text{ softmax}(QK^T))    
\end{equation}
Similar to the original design of \cite{vaswani2017attention},  we divide the number of channels into multiple ‘heads’ and process those in parallel.

In a typical transformer architecture, the self-attention operation is followed by feed-forward and normalization layers. Motivated by \cite{dauphin2017language}, we deploy a gated feed-forward function. The gating mechanism controls the information passed on in the hierarchy. Given the feature map $Y$, a gating function can be generated as $G = \psi(\text{conv}(LN(Y)))$, where $\psi$ is a non-linear function, such as sigmoid, GELU, etc., $LN$ represents a standard layer-normalization operation \cite{vaswani2017attention}. Given $G$, the output of the gated-feedforward layer can be expressed as
\begin{equation}
\label{eq:ff}
Y^{TG} = Y + G \otimes \text{conv}(LN(Y))    
\end{equation}
Ultimately, we will have four different feature representations from four different levels $Y^{TG}_e \in \mathbb{R}^{C*e \times \frac{H}{2^{(e-1)}} \times \frac{H}{2^{(e-1})}}, e \in \{1,2,3,4\}$.
\subsubsection{Low-resolution Local Encoder (LRL):}
We introduce two convolutional branches to compensate for the deficiencies of fine-grained local information in the global transformer branch. In contrast to methods like \cite{Johnston_2020_CVPR}, which embed the self-attention layer into the traditional CNN, our global branch encodes the images independently, which can fully exert the advantages of the transformer. To complement this operation, first, we deploy a ResNet-based low-resolution local branch. We use the first four levels of a pre-trained ResNet18 architecture. Given the input image $I$, it gradually convolves and downsamples the feature maps. At four different levels of the encoder, we will again have feature maps $Y^{LRL}_e \in \mathbb{R}^{C*e \times \frac{H}{2^{(e-1)}} \times \frac{H}{2^{(e-1})}}, e \in \{1,2,3,4\}$. $Y^{LRL}_e$ can be expressed as
\begin{equation}
\label{eq:lrl}
Y^{LRL}_e = \text{conv}(\text{down}(Y^{LRL}_{e-1})) + \text{down}(Y^{LRL}_{e-1})
\end{equation}
The final and deepest feature map of the LRL encoder, i.e., $Y^{LRL}_4$ has the lowest spatial resolution. Thus, it loses much of the finer pixel details but is able to capture the overall scene content better with a large number of feature maps. 
\subsubsection{High-resolution Local Encoder (HRL):}
For the high-resolution counterpart, we maintain the spatial resolution of the feature maps at multiple levels. Specifically, we deploy four parallel streams, where the spatial resolution is fixed for a particular stream. But, different streams have different resolutions. This design is derived from \cite{wang2020deep}. It allows us to generate deep, high-resolution feature maps throughout the entire encoding process.

Let ${(Y0)}^{HRL}_e$ be the input for the $e^{th}$ stream. The output ${Y}^{HRL}_e$ of the $e^{th}$ stream can be expressed as 
\begin{equation}
Y^{HRL}_e = \text{conv}({(Y0)}^{HRL}_e) + {(Y0)}^{HRL}_e
\end{equation}
where ${Y}^{HRL}_e \in \mathbb{R}^{C*e \times \frac{H}{2^{(e-1)}} \times \frac{H}{2^{(e-1})}}$. Unlike Eq. \ref{eq:lrl}, we do not use any downsampling operation inside a stream. Thus, for the higher resolution streams, $Y^{HRL}_e$ is much more deeper and expressive than $Y^{LRL}_e$. It helps in predicting highly accurate local depth boundaries. On the contrary, for lower spatial resolution, $Y^{LRL}_e$ is more robust than $Y^{HRL}_e$ due to the large number of feature maps and captures the overall scene content better.
\subsubsection{Fusion Module:}
\begin{figure}
    \centering
    \includegraphics[width = 0.95\textwidth]{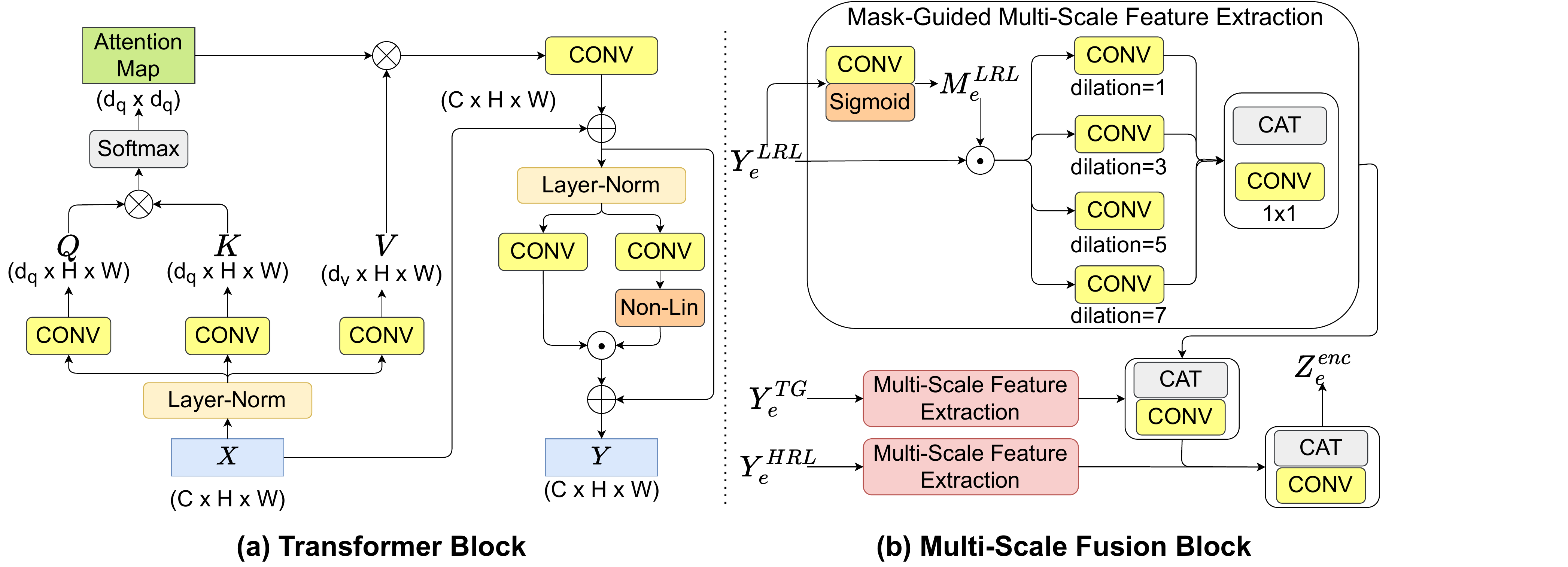}
    \caption{An overview of the transformer block and the multi-scale fusion module.}
    \label{fig:txfusion}
    \vspace{-7mm}
\end{figure}

At the end of the encoder, we will have $Y^{TG}_e, Y^{LRL}_e, Y^{HRL}_e$, for each level  $e \in \{1,2,3,4\}$. To fuse the complementary information from these three branches, we design an atrous convolution-based fusion block (Fig. \ref{fig:txfusion} (b)) that extracts multi-scale information from the feature maps. In most existing works, the final prediction is performed at multiple scales. The coarser outputs usually contain the global structure but may lack intricate local details, which are present in the finer predictions. Motivated by this observation, we propose a multi-convolution-based fusion block that aims to extract and fuse features at multiple scales. We deploy multiple convolution layers with different dilation rates in parallel. Convolution with dilation 1 will better capture the immediate local information, whereas a higher dilation rate will capture the coarser but more global information. We concatenate these feature representations and fuse them using a $1 \times 1$ convolutional layer. Our experimental results demonstrate that such operations in the feature domain significantly help the depth estimation network capture a better representation of the scene and boost the accuracy.

We use four parallel convolution layers with different dilation rates (1, 3, 5, 7) to extract information at multiple scales. Formally, for input $Y \in \mathbb{R}^{C \times H \times W}$, multiple atrous convolutions are applied over the input feature map as follows:
\begin{equation}
\label{eq:atr1}
    Y^r[j] = \sum_k Y[j + r \cdot k]W[k]    
\end{equation}
where $k$ is the size of the filter $W$ and the dilation rate $r$ determines the stride. Standard convolution is a special case in which rate $r = 1$. Next, we concatenate these outputs and pass through a $1 \times 1$ convolution layer to maintain the number of feature maps. The operation can be expressed as 
\begin{equation}
\label{eq:atr2}
    Z = \text{conv}_{1 \times 1}(\text{concat}(Y^1, Y^3, Y^5, Y^7))
\end{equation}
It allows the network to capture multi-scale abstractions of the scene.
\newline Before extracting multi-scale information from $Y^{TG}_e, Y^{LRL}_e, Y^{HRL}_e$ using Eq. \ref{eq:atr2}, we emphasize the most useful feature locations for the local convolutional branches using a spatial masking technique. Given, $Y^{LRL}_e \in \mathbb{R}^{\hat{C} \times \hat{H} \times \hat{W}}$, we first generate a spatial mask $M^{LRL} \in \mathbb{R}^{\hat{C} \times \hat{H} \times \hat{W}}$. Next, we elementwise multiply the feature map $Y^{LRL}_e$ with the corresponding spatial mask as
\begin{align*}
\label{eq:m_lrl}
    M^{LRL} = \text{sigmoid}(\text{conv}(Y^{LRL}_e) \\
    Y'^{LRL}_e = Y^{LRL}_e \odot M^{LRL}
\end{align*}
This process allows us to enhance the crucial and helpful features from the local branches while suppressing the redundant ones. The same is applied to generate $Y'^{HRL}_e$ as well. Next, we pass each of these enhanced feature maps ($Y'^{LRL}_e, Y'^{HRL}_e$ and $Y^{TG}_e$) through multiple atrous-convolution layers (Eqs. \ref{eq:atr1} and \ref{eq:atr2}) to generate the updated feature maps $Z^{LRL}_e, Z^{HRL}_e, Z^{TG}_e$ with multi-scale information. Next, we concatenate a pair of feature sets for a particular level and pass through an $1 \times 1$ convolutional layer to maintain the number of channels. This process is repeated twice to fuse information from three branches. Formally, the final output of encoder for level $e$ can be expressed as
\begin{equation}
    Z_e^{enc} = \text{conv}_{1 \times 1}(\text{concat}(\text{conv}_{1 \times 1}(\text{concat}(Z^{TG}_e, Z^{LRL}_e)), Z^{HRL}_e)) 
\end{equation}

Note that, information from all three branches can be concatenated and fused at one go, but it increases the computational load due to high number of feature channels. 
\subsubsection{Decoder}
We aim to extract a robust set of feature representations in the encoder. Information from different branches is merged in the proposed fusion block. Finally, the input to the decoder is $Z_4^{enc}$. This is further refined in the first decoder level and upsampled at the end. We resort to a purely transformer-based decoder as it excels in modeling the dependencies between different feature maps and locations. The core operations are the same as Eqs. \ref{eq:att} and \ref{eq:ff}. The upsampled decoder feature is concatenated with the encoder feature of the same level $Z_3^{enc}$, constituting the input to the next decoder level. This process is repeated for all levels. At the end, we pass the feature map through a convolution followed by a sigmoid layer to estimate the final disparity map. Note that, while training, we predict the disparity map at every decoder level following the multi-scale prediction approach of \cite{monodepth2}. Thus, we simply add one convolution and sigmoid layer at every decoder level for coarser predictions. During inference, we discard these layers and only take the full-scale output at the end. 

\subsection{Pose Estimation Network}
The input to the pose estimation network is the target view $I_0$ concatenated with one of the source views $I_i$, $i \in \{-1,+1\}$. It predicts the relative pose $T_{0 \rightarrow i}$ between the two views.
For pose estimation, we follow \cite{monodepth2} and use a ResNet18-based network. It consists of a series of residual convolution and downsampling blocks derived from ResNet18. In the end, we use a $1 \times 1$ convolutional layer and a 6-channel spatial average pooling operation. The output is the set of 6 DoF transformations between $T_{0 \rightarrow i}$ (corresponding to 3 Euler angles and 3-D translation). The translation is parametrized in Euclidean coordinates $\{x, y, z\}$ and rotation uses Euler angles $\{\alpha, \beta, \gamma\}$. This process is repeated for each source view producing independent transformations. 

\subsection{Training Strategies}
Following \cite{monodepth2}, we also use the assumption that the world is static and the view change is only caused by moving the camera. So, we should be able to synthesize the target view $I_0$ given the depth map of the scene, pixel intensities of the source view, relative pose, and the camera intrinsics ($K$). Usually, $K$ is known for a particular dataset. 
Given the relative pose $T_{0 \rightarrow i}$ and the depth map $D_0$ predicted by the pose and depth estimation networks, respectively, the view synthesis operation can be expressed as
\begin{equation}
    I'_{i \rightarrow 0} = I_i <\text{proj}_{cam}(\text{proj}_{world}(I_0, D_0, T_{0 \rightarrow i}), K)>
\end{equation}
where $\text{proj}_{world}$ represents the inverse projection of image points to 3D world points, $\text{proj}_{cam}$ represents the projection of 3D coordinates to image points, $<>$ represents the sampling operation. We use the differentiable bilinear sampling mechanism of \cite{jaderberg2015spatial} that linearly interpolates the values of the 4-pixel neighbors (top-left, top-right, bottom-left, and bottom-right) to calculate $I'_{i \rightarrow 0}$.

Next, we utilize two main regularizing function for training the network. We employ photometric loss to calculate the difference between $I_0$ and $I'_{i \rightarrow 0}$. Following \cite{monodepth2,Johnston_2020_CVPR}, the loss function can be  defined as
\begin{equation}
    R_p^i = \frac{\alpha}{2} (1 - \text{SSIM}(I_0, I'_{i \rightarrow 0})) + (1-\alpha)||I_0 - I'_{i \rightarrow 0} ||_1
\end{equation}
where SSIM represents Structural Similarity \cite{wang2004image} and $\alpha = 0.85$. Moreover, to promote the smoothness of the generated depth map, we use the widely used edge-aware smoothness during training 
\begin{equation}
    R_s = |\partial_x d_0|e^{-|\partial_x I_0|} + |\partial_y d_0|e^{-|\partial_y I_0|}
\end{equation}
We also utilize the minimum photometric error, masking of the stationary pixels and multi-scale losses introduced in \cite{monodepth2}. The final loss is computed as the weighted sum of the minimum reprojection loss and the smoothness term
\begin{equation}
    R_{final} = \min_i R_p^i + \beta R_s
\end{equation}
\section{Experiments}
\subsection{Dataset Details}
The KITTI dataset \cite{6248074} is an accepted standard dataset for the task of Monocular Depth Estimation. It's Eigen Split \cite{NIPS2014_7bccfde7} for Monocular Depth Estimation comprises 22,600 training images, 888 validation images, and 697 test images. Our model uses images of dimensions $640 \times 192 \times 3$ for training and those of dimensions $1280 \times 384 \times 3$ for evaluation. We apply the same preprocessing techniques as \cite{monodepth2}. 
\subsection{Training and Implementation Details}
We build upon the standard framework of Monodepth2 \cite{monodepth2}. All presented models were trained on the KITTI \cite{6248074} dataset. Our model was trained using 4 NVIDIA RTX 3090 GPUs with a batch size of 16 for 20 epochs using Adam \cite{https://doi.org/10.48550/arxiv.1412.6980} optimizer at an initial learning rate of $10^{-4}$ with a geometric progression based decay schedule. In the following sections, we present quantitative and qualitative results to back our findings, along with several ablation studies to corroborate the importance of each element of our proposed method and the synergy between them.

\begin{center}
\begin{table}[!t]
    \caption{Quantitaive Results for our model's performance with respect to SOTA. The best performing entries have been highlighted in bold. The second best entries have been underlined. Methods indicated with $*$ and $\dagger$ require semantic data and GT depth map, respectively. Our model performs best on five and second best on one of the seven metrics. ( $\downarrow$: lower is better, $\uparrow$: higher is better.)}
    \label{tab:quant}
    \resizebox{12cm}{!}{  
    \begin{tabular}{|c|c||c|c|c|c||c|c|c|} 
        \hline
        \textbf{Data} & \textbf{Method} & \textbf{Abs Rel} $\downarrow$ & \textbf{Sq Rel} $\downarrow$ & \textbf{RMSE} $\downarrow$ & \textbf{RMSE log} $\downarrow$ & \tiny $\delta$ $<$ 1.25 $\uparrow$ & \tiny $\delta$ $<$ $1.25^{2}$ $\uparrow$ & \tiny $\delta$ $<$ $1.25^{3}$ $\uparrow$ \\
        \hline  
        \multirow{20}{*}{\cite{6248074}} & Zhou \textit{et al.} \cite{Zhou_2017_CVPR} & 0.208 & 1.768 & 6.856 & 0.283 & 0.678 & 0.885 & 0.957\\
        & GeoNet \cite{Qi_2018_CVPR}$\dagger$ & 0.164 & 1.303 & 6.090 & 0.247 & 0.765 & 0.919 & 0.968\\
        & Struct2Depth \cite{DBLP:journals/corr/abs-1811-06152}{*} & 0.141 & 1.026 & 5.291 & 0.215 & 0.816 & 0.945 & 0.979\\
        & Mahjourian \cite{Mahjourian_2018_CVPR} & 0.163 & 1.240 & 6.220 & 0.250 & 0.762 & 0.916 & 0.968\\
        & CC \cite{DBLP:journals/corr/abs-1805-09806} & 0.140 & 1.070 & 5.326 & 0.217 & 0.826 & 0.941 & 0.975\\
        & DF-Net \cite{Zou_2018_ECCV} & 0.150 & 1.124 & 5.507 & 0.223 & 0.806 & 0.933 & 0.973\\
        & Li \textit{et al.} \cite{Li_2019_ICCV} & 0.150 & 1.127 & 5.564 & 0.229 & 0.823 & 0.936 & 0.974\\
        & Pilzer \textit{et al.} \cite{Pilzer_2019_CVPR} & 0.142 & 1.231 & 5.785 & 0.239 & 0.795 & 0.924 & 0.968\\
        & $EPC++$ \cite{8769907} & 0.141 & 1.029 & 5.350 & 0.216 & 0.816 & 0.941 & 0.976\\
        & Bian \textit{et al.} \cite{NEURIPS2019_6364d3f0} & 0.137 & 1.089 & 5.439 & 0.217 & 0.830 & 0.942 & 0.975\\
        & GLNet \cite{Chen_2019_ICCV} & 0.135 & 1.070 & 5.230 & 0.210 & 0.841 & 0.948 & 0.980\\
        & Gordon \textit{et al.} \cite{Gordon_2019_ICCV}{*} & 0.128 & 0.959 & 5.230 & 0.212 & 0.845 & 0.947 & 0.976\\
        & Monodepth2 \cite{monodepth2} & 0.115 & 0.882 & 4.701 & 0.19 & 0.879 & 0.961 & \underline{0.982} \\
        & Packnet-SfM \cite{Guizilini_2020_CVPR} & \underline{0.107} & 0.802 & \underline{4.538} & \underline{0.186} & \textbf{0.889} & \underline{0.962} & 0.981\\
        & Semantics \cite{Tosi_2020_CVPR}{*} & 0.126 & 0.835 & 4.937 & 0.199 & 0.844 & 0.953 & \underline{0.982}\\
        & SGDepth \cite{klingner2020selfsupervised} & 0.117 & 0.907 & 4.844 & 0.196 & 0.875 & 0.958 & 0.98 \\
        & Jonston \textit{et al.} \cite{Johnston_2020_CVPR} & \textbf{0.106} & 0.861 & 4.699 & \textbf{0.185} & \textbf{0.889} & \underline{0.962} & \underline{0.982}\\
        & Gao \textit{et al.} \cite{DBLP:journals/corr/abs-2011-09369} & 0.112 & 0.866 & 4.693 & 0.189 & \underline{0.881} & 0.961 & \textbf{0.983} \\
        & Lee \textit{et al.} \cite{Lee_Im_Lin_Kweon_2021} & 0.112 & \underline{0.777} & 4.772 & 0.191 & 0.872 & 0.959 & \underline{0.982}\\
        & \textbf{Ours} & 0.112 & \textbf{0.75} & \textbf{4.528} & 0.187 & \underline{0.881} & \textbf{0.963} & \textbf{0.983} \\
        \hline
        \multirow{3}{*}{\cite{NIPS2005_17d8da81}} & Johnston \textit{et al.} \cite{Johnston_2020_CVPR} & \underline{0.711} &	\underline{15.868} & \underline{27.934} & \underline{1.978} & \textbf{0.115} &	\underline{0.216} &	\underline{0.312}\\
        & Packnet-SfM \cite{Guizilini_2020_CVPR} & 0.725 & 17.102 & 28.352 & 	2.037 &	0.096 &	0.194 &	0.287\\
        & \textbf{Ours} & \textbf{0.601} & \textbf{14.391} & \textbf{26.016} &	\textbf{1.519} & \underline{0.104} & \textbf{0.227}	& \textbf{0.357} \\
        \hline
        \multirow{3}{*}{\cite{Xian_2020_CVPR}} & Jonston \textit{et al.} \cite{Johnston_2020_CVPR} & \underline{0.393} &	\underline{1.839} & \underline{1.102} & \underline{0.274} & \underline{0.115} &	\underline{ 0.199} &	\underline{0.254}\\
        & Packnet-SfM \cite{Guizilini_2020_CVPR} & 1.304 & 6.04 & 3.861& 1.150 &	\textbf{0.121} &	\textbf{0.241} & \textbf{0.356}\\
        & \textbf{Ours} & \textbf{0.188} & \textbf{0.512} & \textbf{0.972} &	\textbf{0.262} & 0.078 & 0.160	& 0.238 \\
        \hline
    \end{tabular}}
\end{table}
\vspace{-6mm}
\end{center}
\subsection{Quantitative Analysis}
Keeping in line with standard practice, we evaluate the efficacy of our model using Absolute Relative Error (Abs Rel), Square Relative Error (Sq Rel), Root Mean Squared Error (RMSE), Log RMSE, and the standard accuracy thresholds viz. $\delta$ $<$ 1.25, $\delta$ $<$ $1.25^{2}$, and $\delta$ $<$ $1.25^{3}$. Amongst these, RMSE is the most stringent error measure since it is the RMS value of the absolute pixel-wise error observed between a predicted depth map and the ground truth, whereas others except Log RMSE are relative to a certain degree. 
In Table \ref{tab:quant}, we have reported the quantitative scores of our model and existing supervised, semantic-guided, and unsupervised depth estimation networks. Our model achieves SOTA performance on KITTI Eigen split \cite{NIPS2014_7bccfde7} with respect to RMSE and also beats existing methods on most other metrics. For completeness, we also evaluate on a subset of images from Make3D \cite{NIPS2005_17d8da81} \cite{4531745} and HR-WSI \cite{Xian_2020_CVPR} datasets. We compare our approach with two recent SOTA works - \cite{Johnston_2020_CVPR} and \cite{Guizilini_2020_CVPR} with open-source implementation. All the models were trained on KITTI and then tested on a subset of images from \cite{NIPS2005_17d8da81} and \cite{Xian_2020_CVPR}.  As can be observed in Table \ref{tab:quant}, our approach achieves superior performance on diverse range of scenes from other datasets, as well.      

\subsection{Qualitative Analysis}
\begin{figure*}[t]
\setlength{\tabcolsep}{1pt}
\scriptsize
\centering
\resizebox{1\textwidth}{!}{%
\begin{tabular}{cccc}
\includegraphics[width = \textwidth, height = 0.3\textwidth]{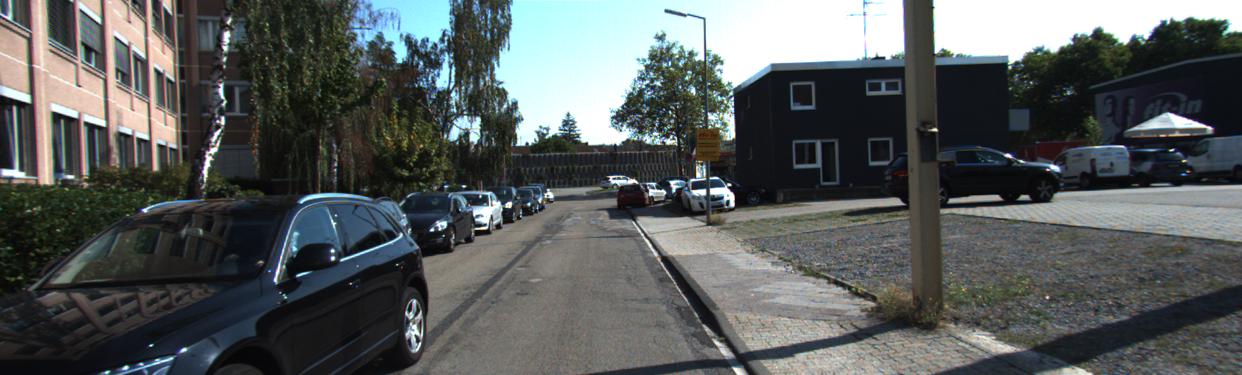} & \includegraphics[width = \textwidth, height = 0.3\textwidth]{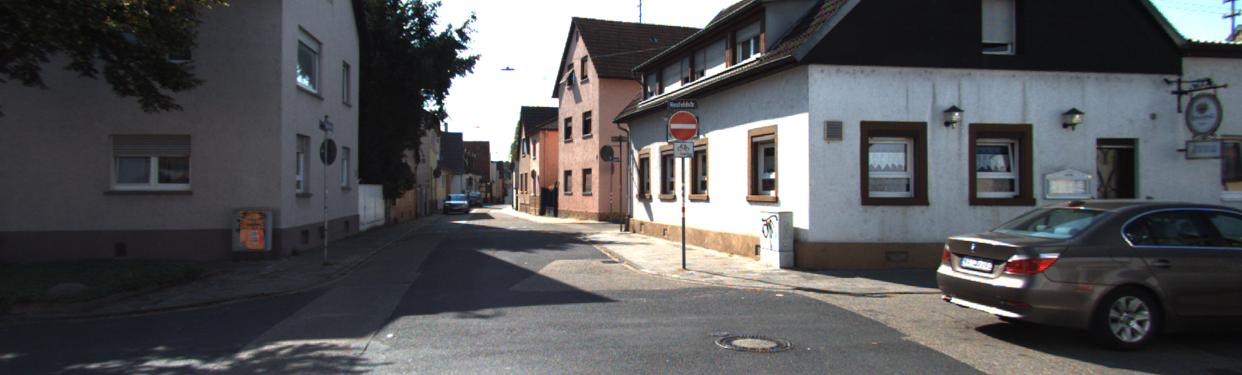} & \includegraphics[width = \textwidth, height = 0.3\textwidth]{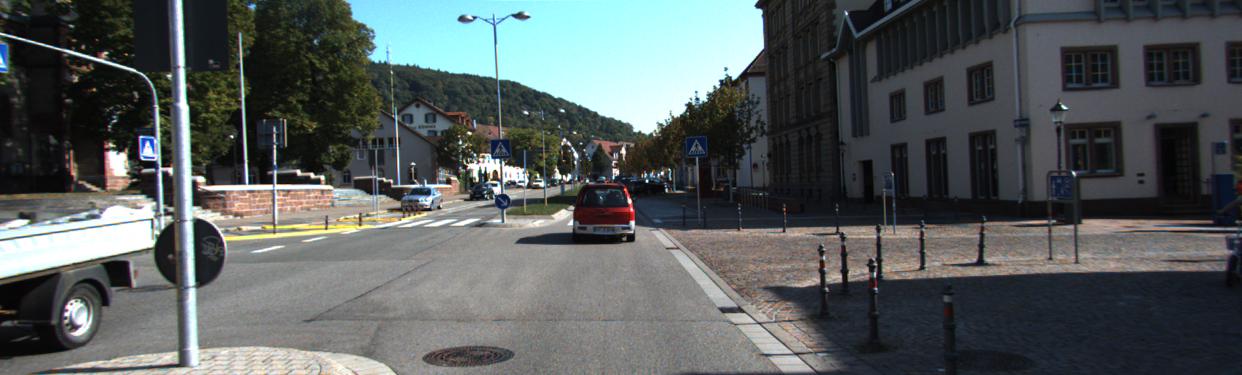} & \includegraphics[width = \textwidth, height = 0.3\textwidth]{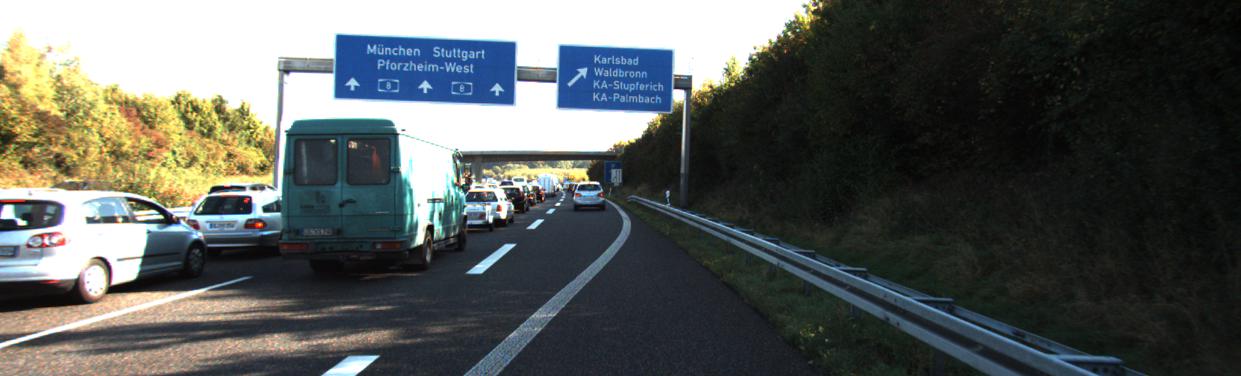} \\
 \includegraphics[width = \textwidth, height = 0.3\textwidth]{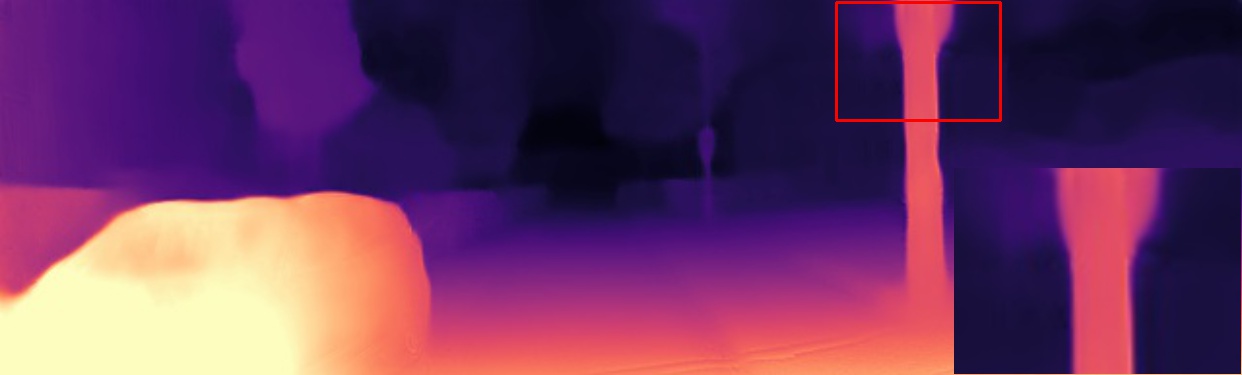} & \includegraphics[width = \textwidth, height = 0.3\textwidth]{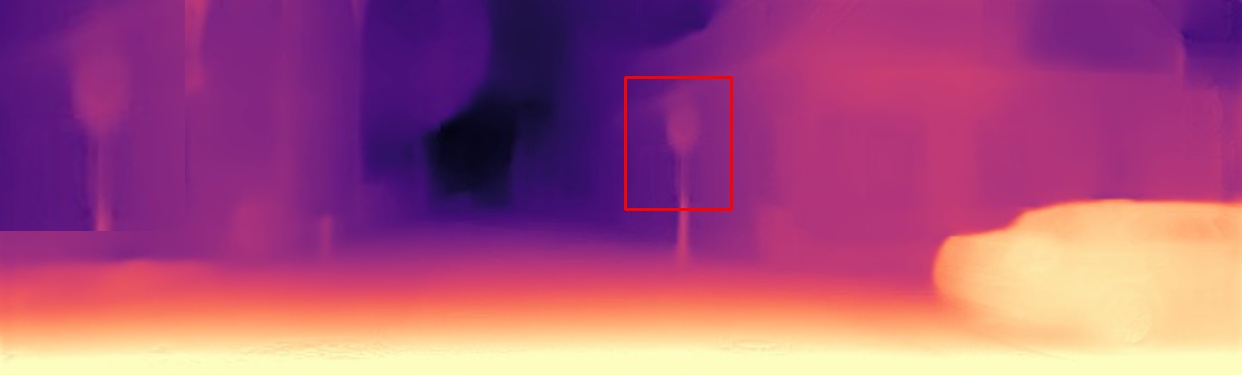} & \includegraphics[width = \textwidth, height = 0.3\textwidth]{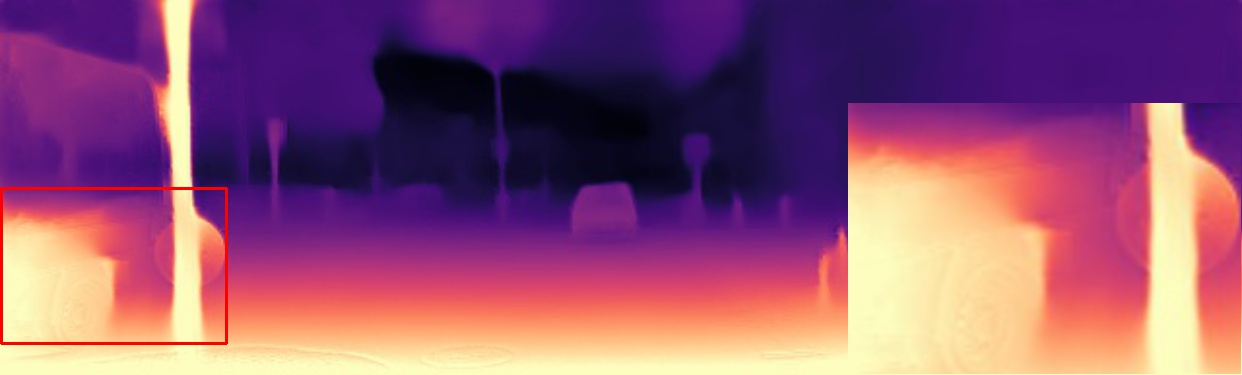} & \includegraphics[width = \textwidth, height = 0.3\textwidth]{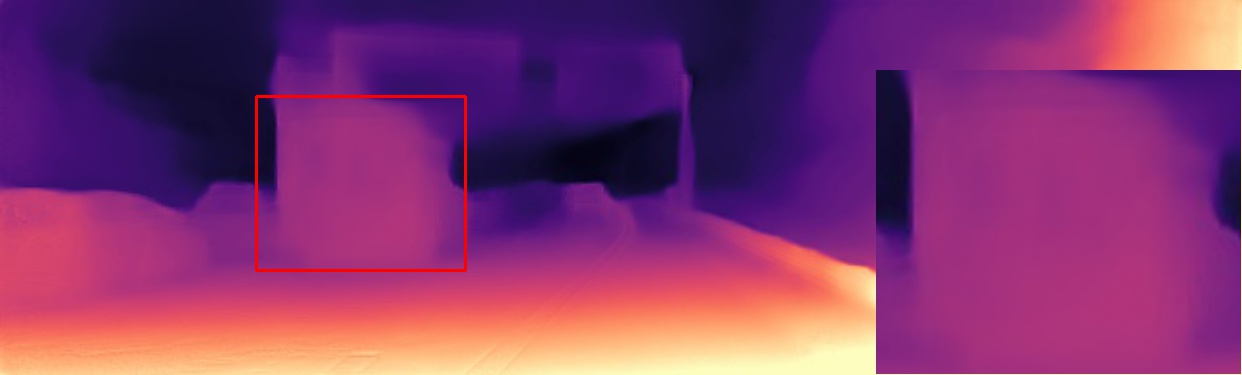} \\
\includegraphics[width = \textwidth, height = 0.3\textwidth]{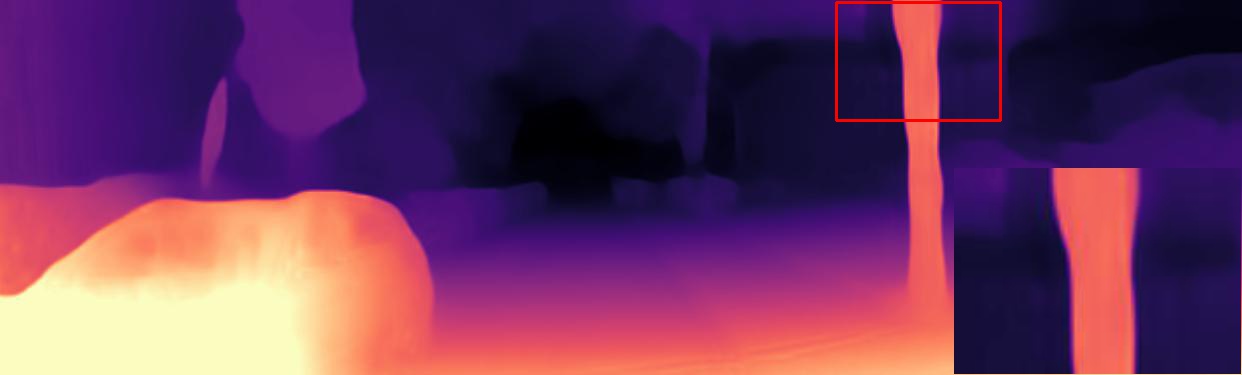} & \includegraphics[width = \textwidth, height = 0.3\textwidth]{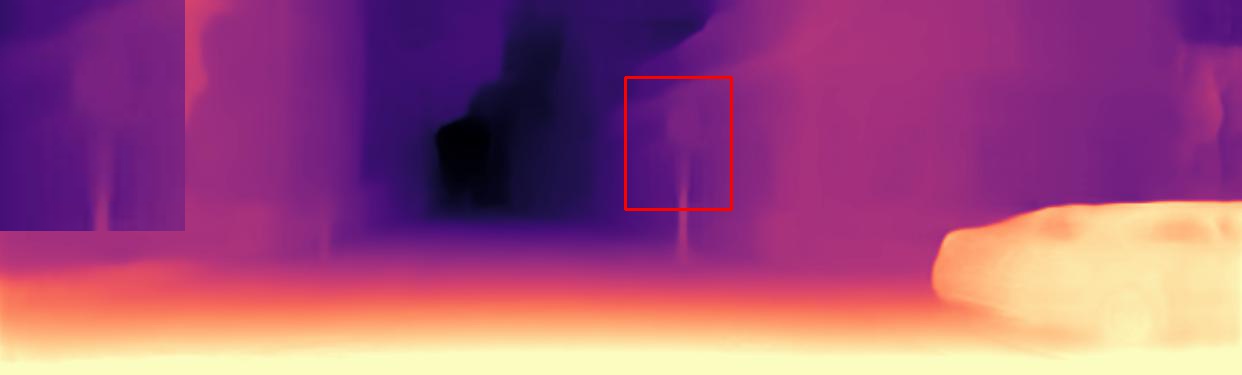} & \includegraphics[width = \textwidth, height = 0.3\textwidth]{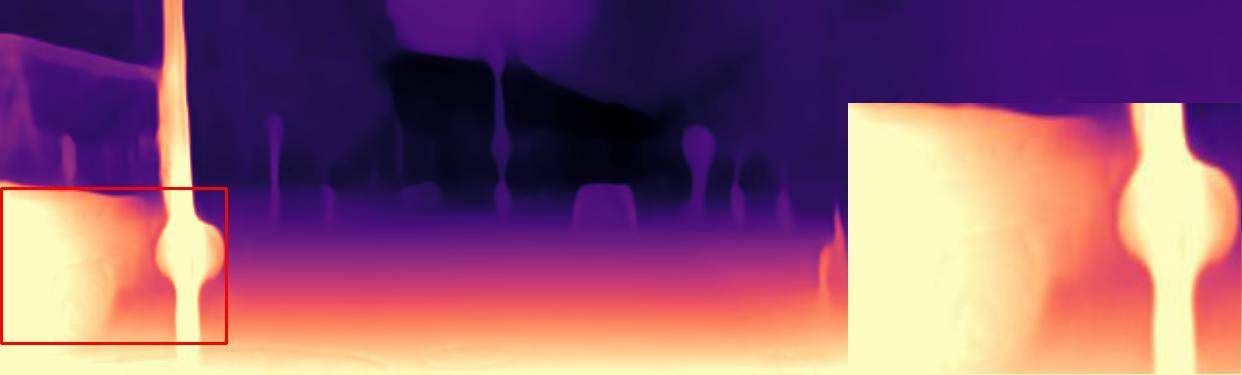} & \includegraphics[width = \textwidth, height = 0.3\textwidth]{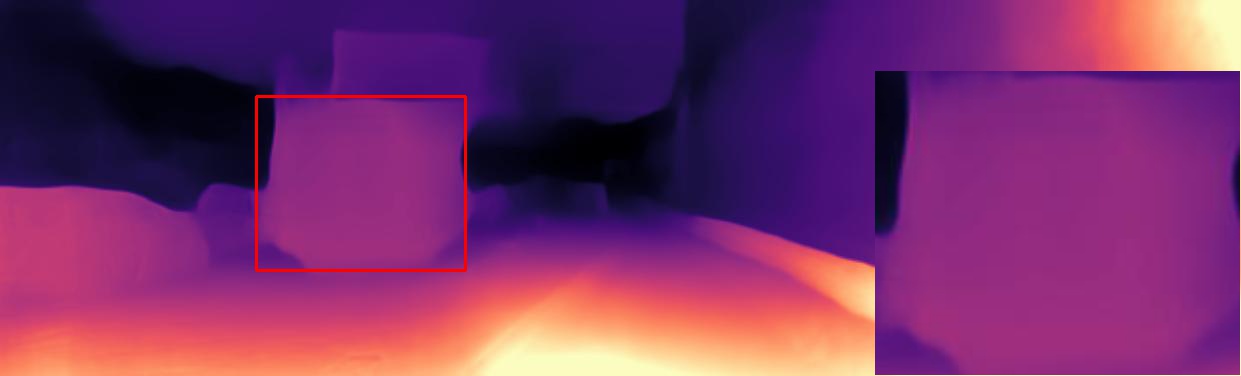} \\
\includegraphics[width = \textwidth, height = 0.3\textwidth]{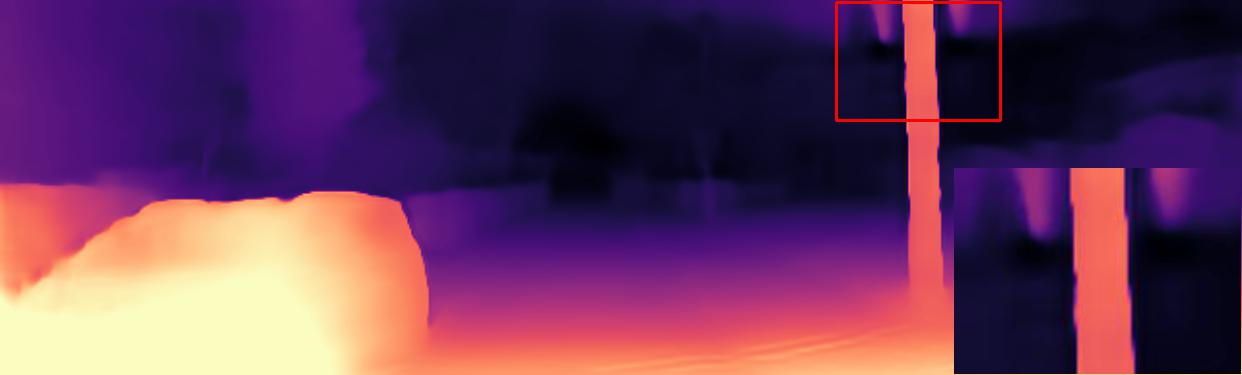} & \includegraphics[width = \textwidth, height = 0.3\textwidth]{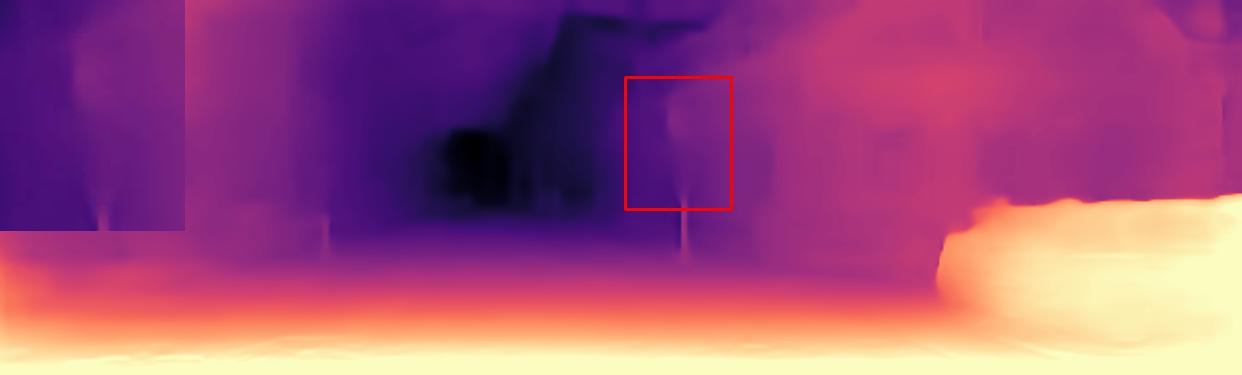} & \includegraphics[width = \textwidth, height = 0.3\textwidth]{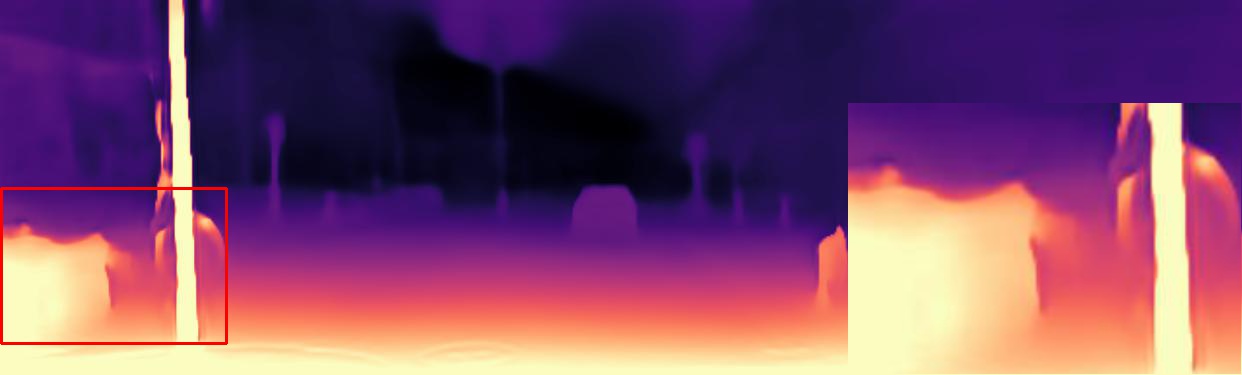} & \includegraphics[width = \textwidth, height = 0.3\textwidth]{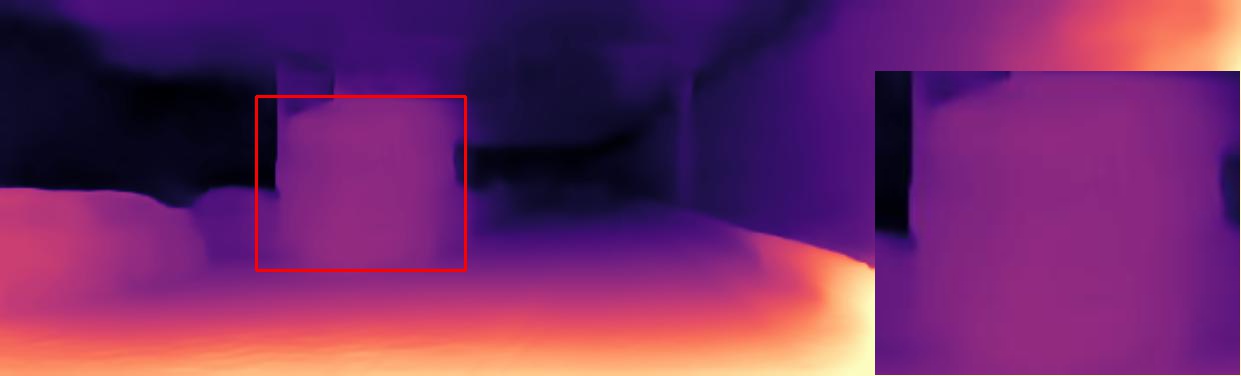} \\
\includegraphics[width = \textwidth, height = 0.3\textwidth]{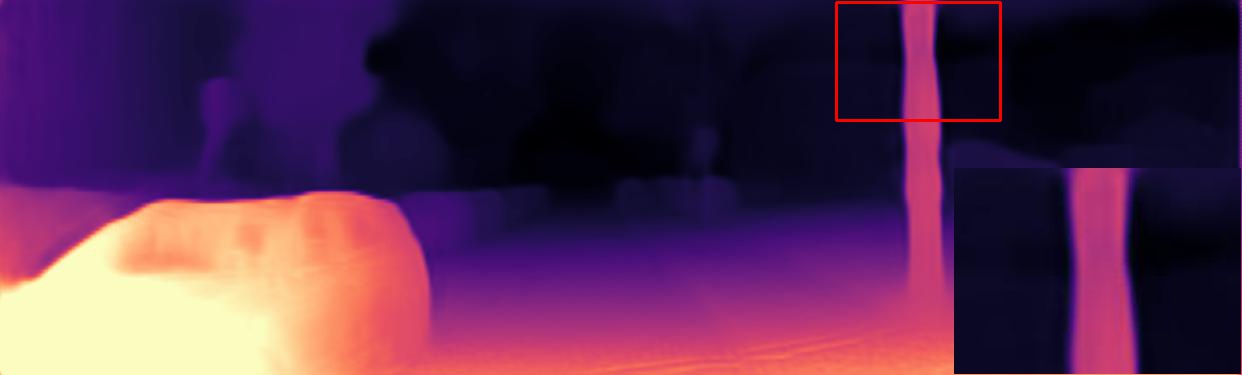} & \includegraphics[width = \textwidth, height = 0.3\textwidth]{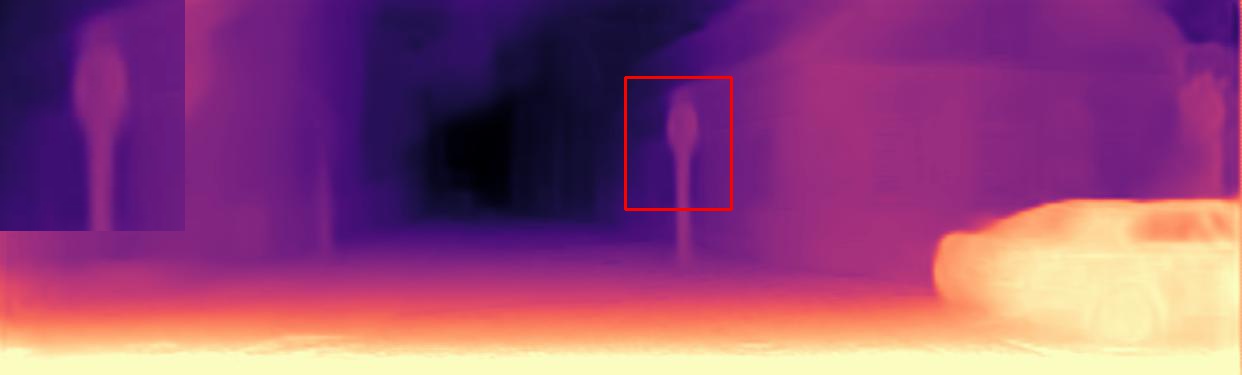} & \includegraphics[width = \textwidth, height = 0.3\textwidth]{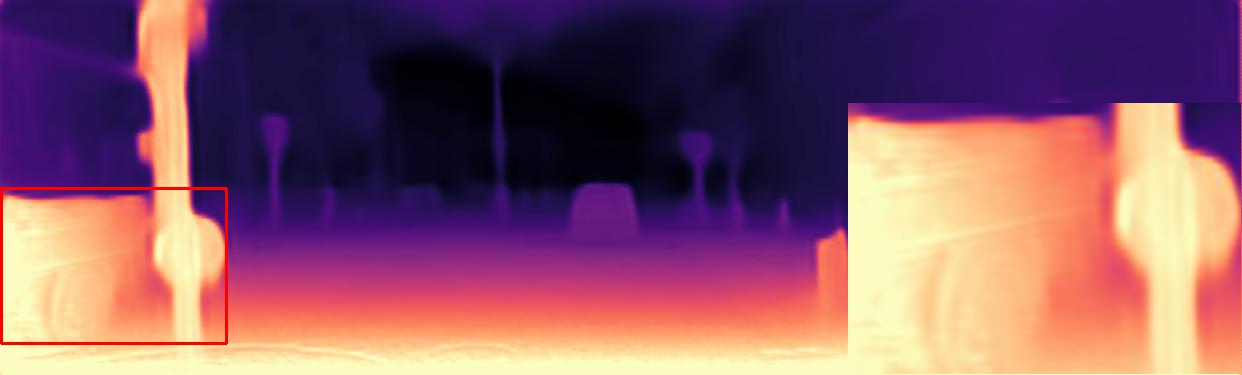} & \includegraphics[width = \textwidth, height = 0.3\textwidth]{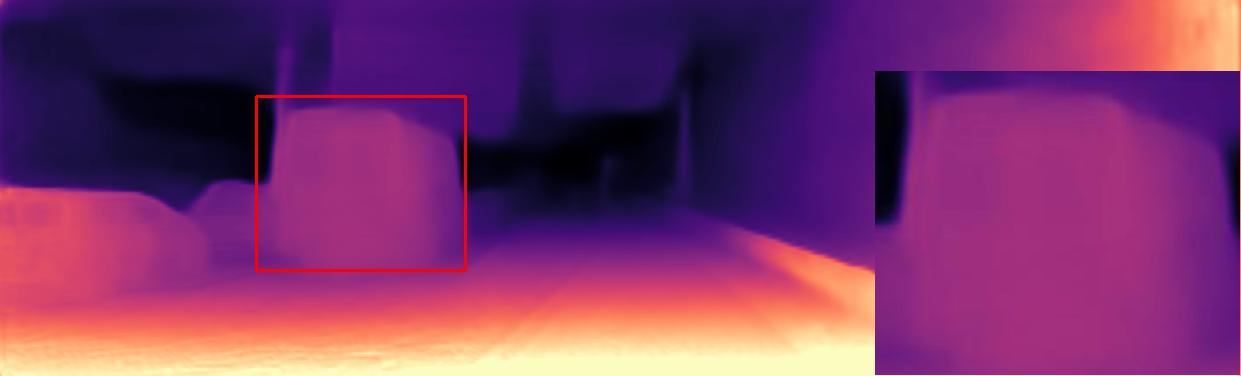} 
\end{tabular}%
}

\caption{Qualitative comparisons on selected test images from the  Eigen split \cite{NIPS2014_7bccfde7} benchmark of the KITTI dataset \cite{6248074}. From top: Input, Packnet-SfM \cite{Guizilini_2020_CVPR}, Monodepth2 \cite{monodepth2}, Jonston \textit{et al.} \cite{Johnston_2020_CVPR} and our outpus.}
\label{fig:qual}
\vspace{-2mm}
\end{figure*}

\begin{figure*}[!h]
\setlength{\tabcolsep}{1pt}
\scriptsize
\centering
\resizebox{1\textwidth}{!}{%
\begin{tabular}{ccccc}
\includegraphics[width = \textwidth, height = 0.62\textwidth]{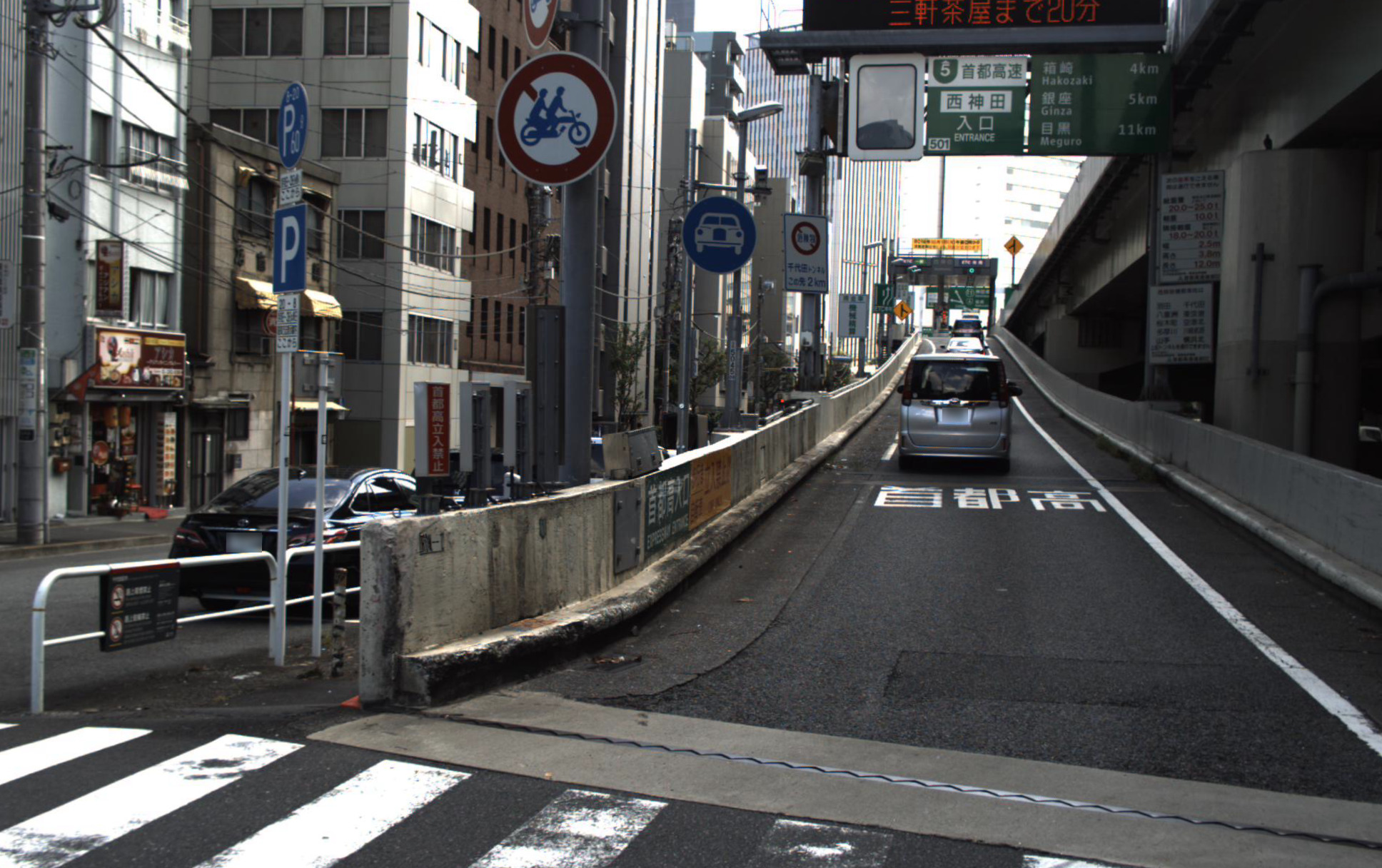} & \includegraphics[width = \textwidth, height = 0.62\textwidth]{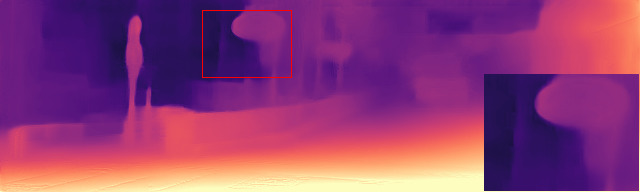} & \includegraphics[width = \textwidth, height = 0.62\textwidth]{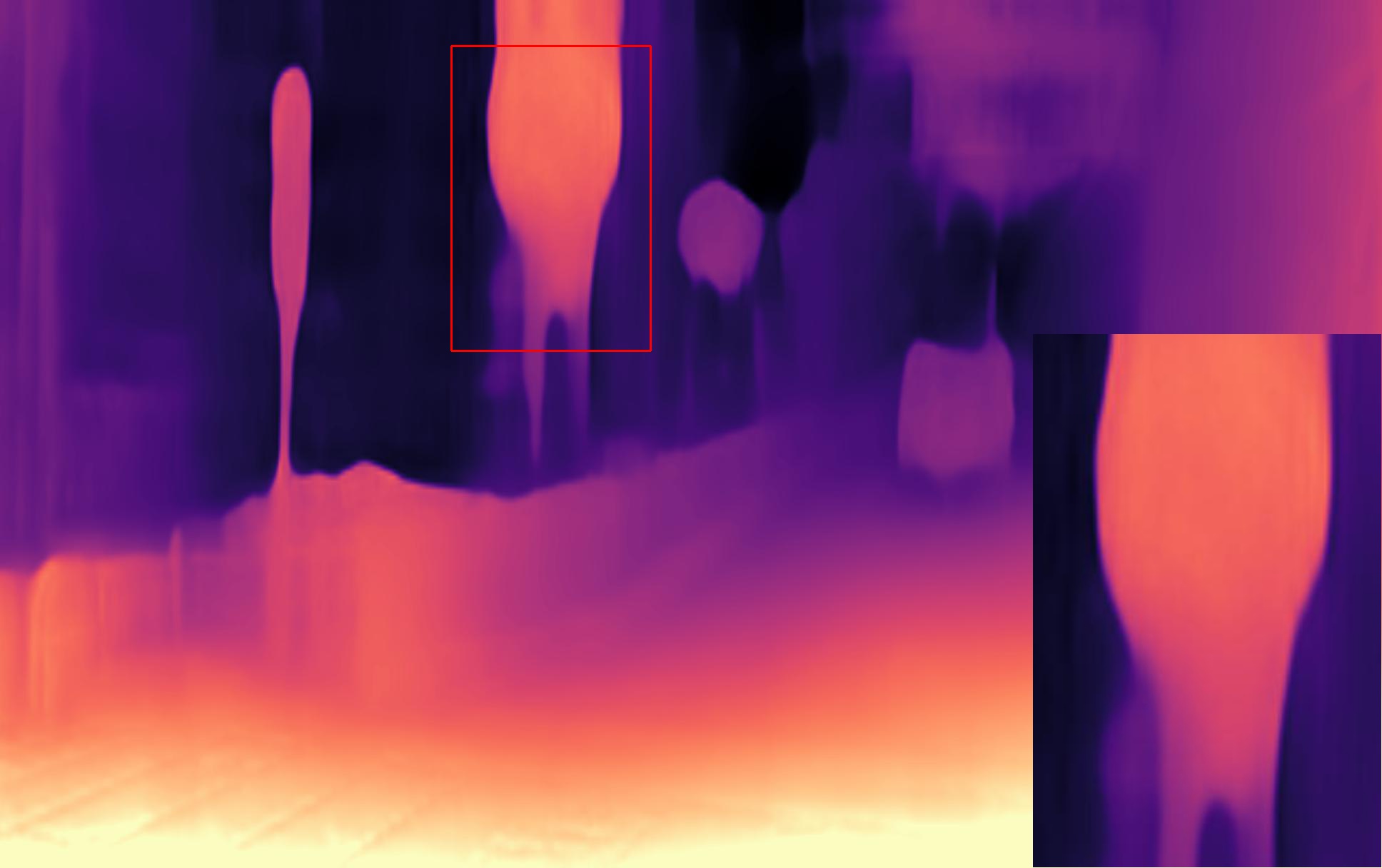} & \includegraphics[width = \textwidth, height = 0.62\textwidth]{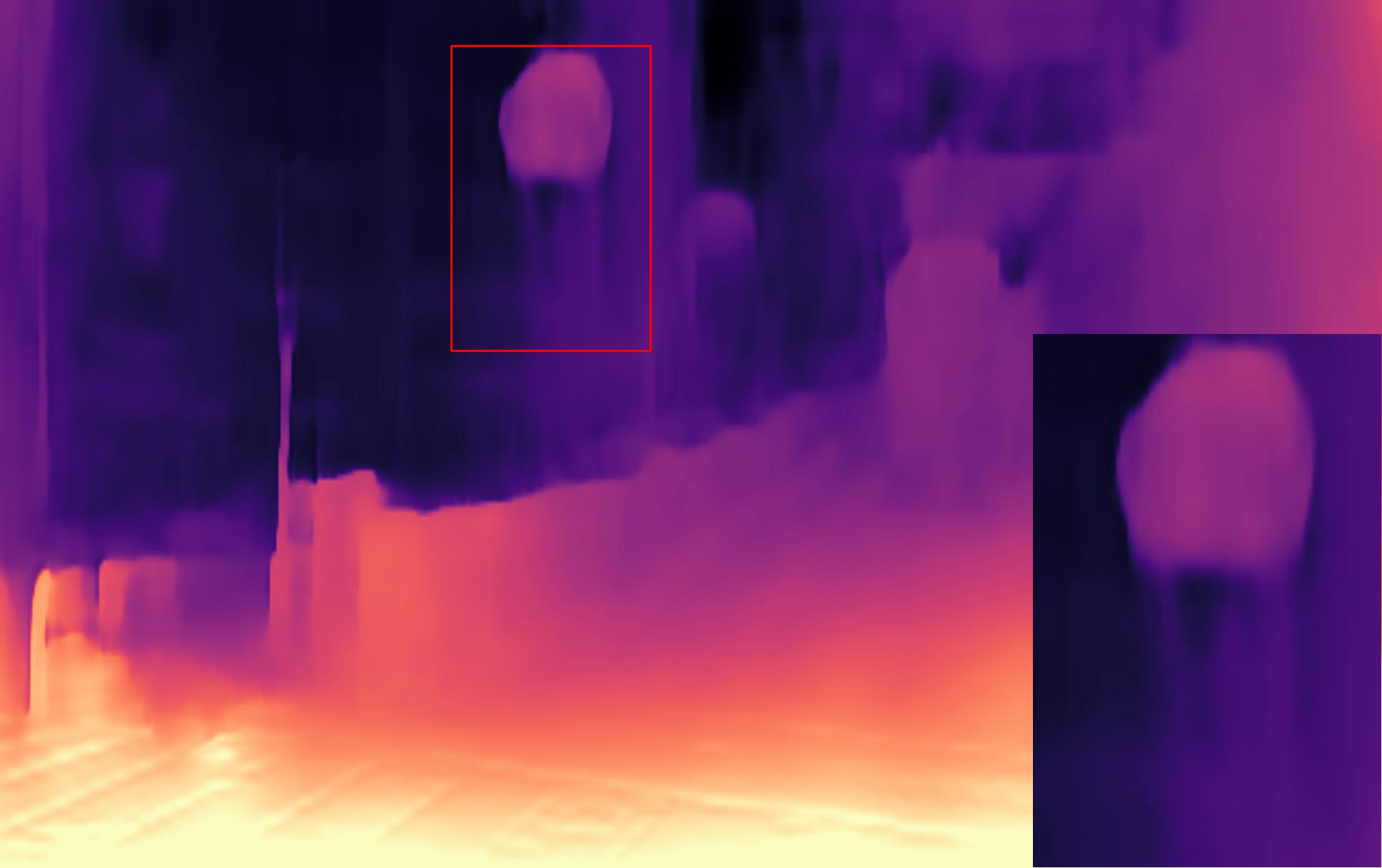} & \includegraphics[width = \textwidth, height = 0.62\textwidth]{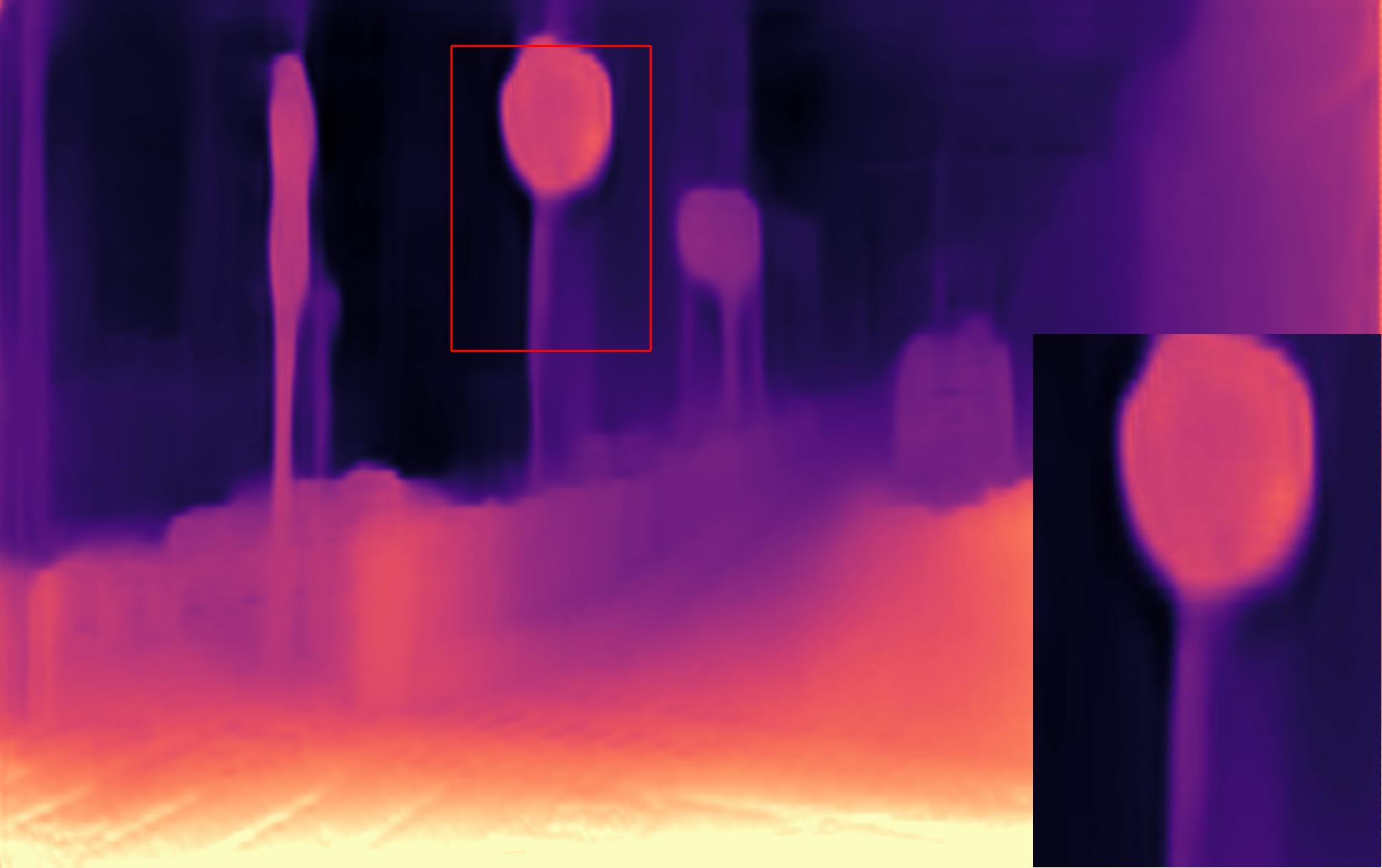}\\ 
\includegraphics[width = \textwidth, height = 0.62\textwidth]{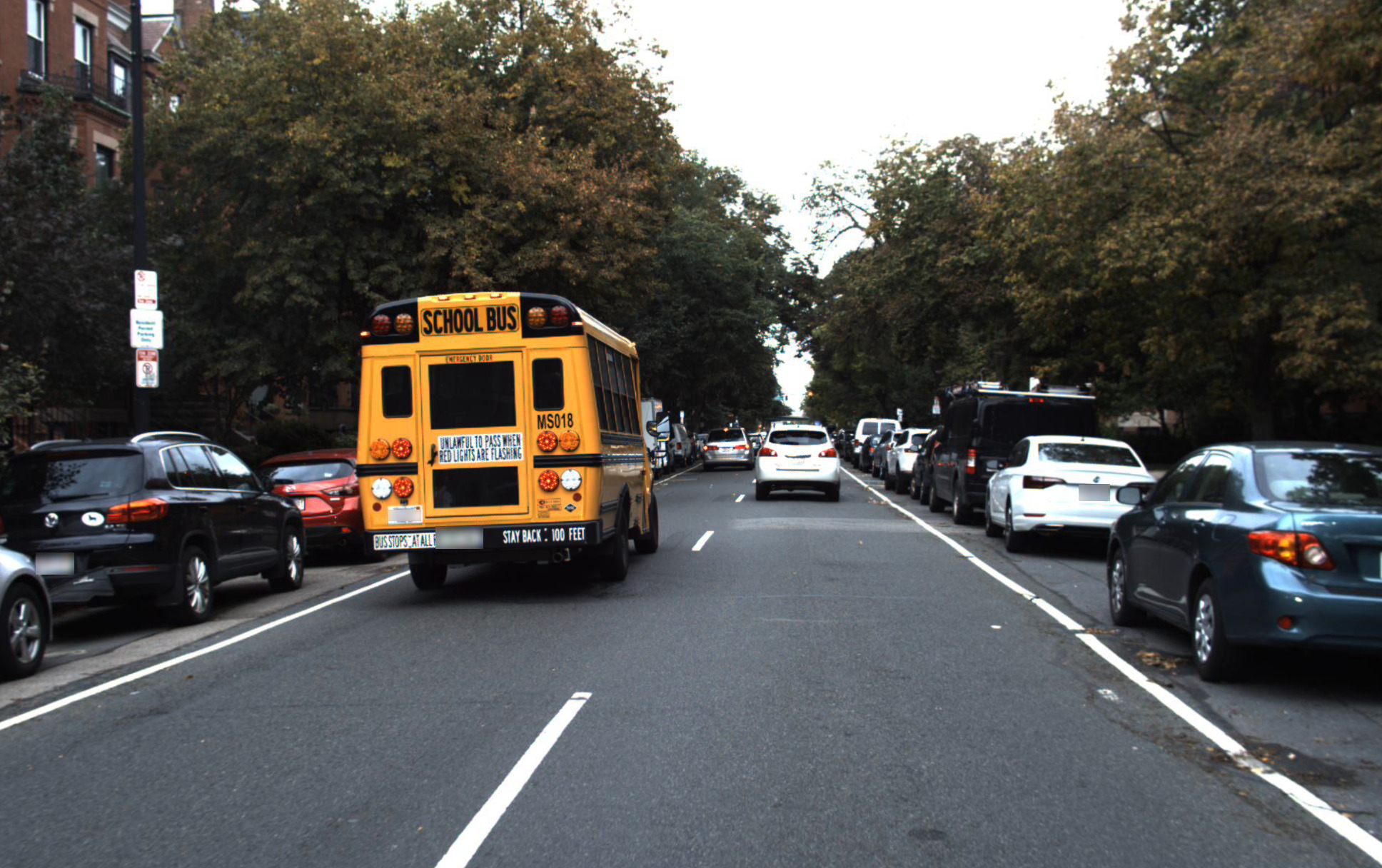} & \includegraphics[width = \textwidth, height = 0.62\textwidth]{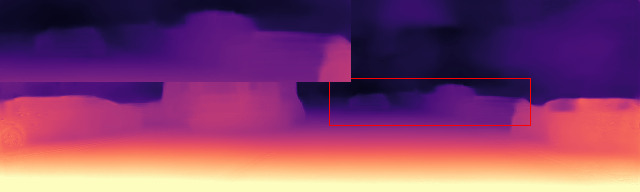} & \includegraphics[width = \textwidth, height = 0.62\textwidth]{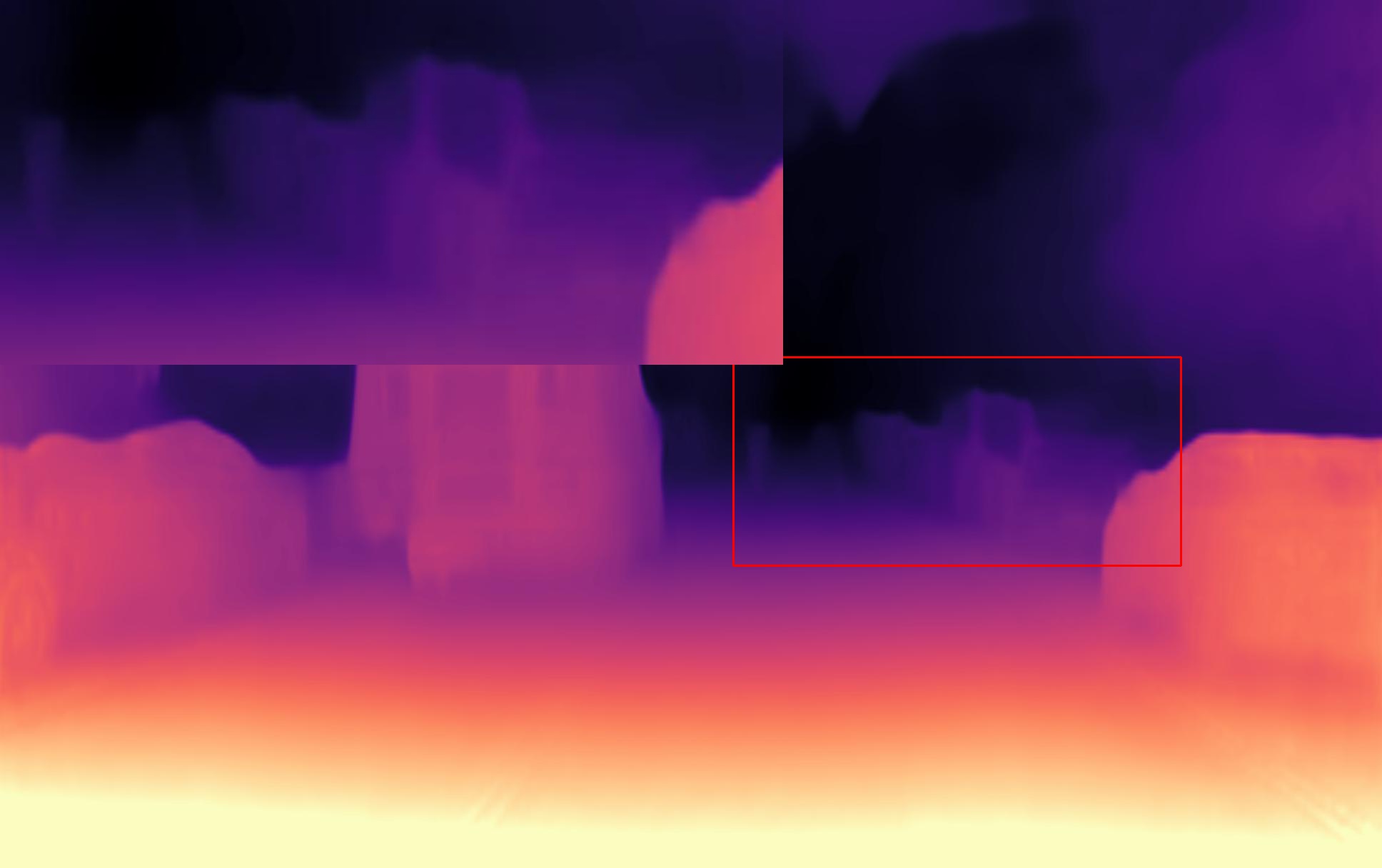} & \includegraphics[width = \textwidth, height = 0.62\textwidth]{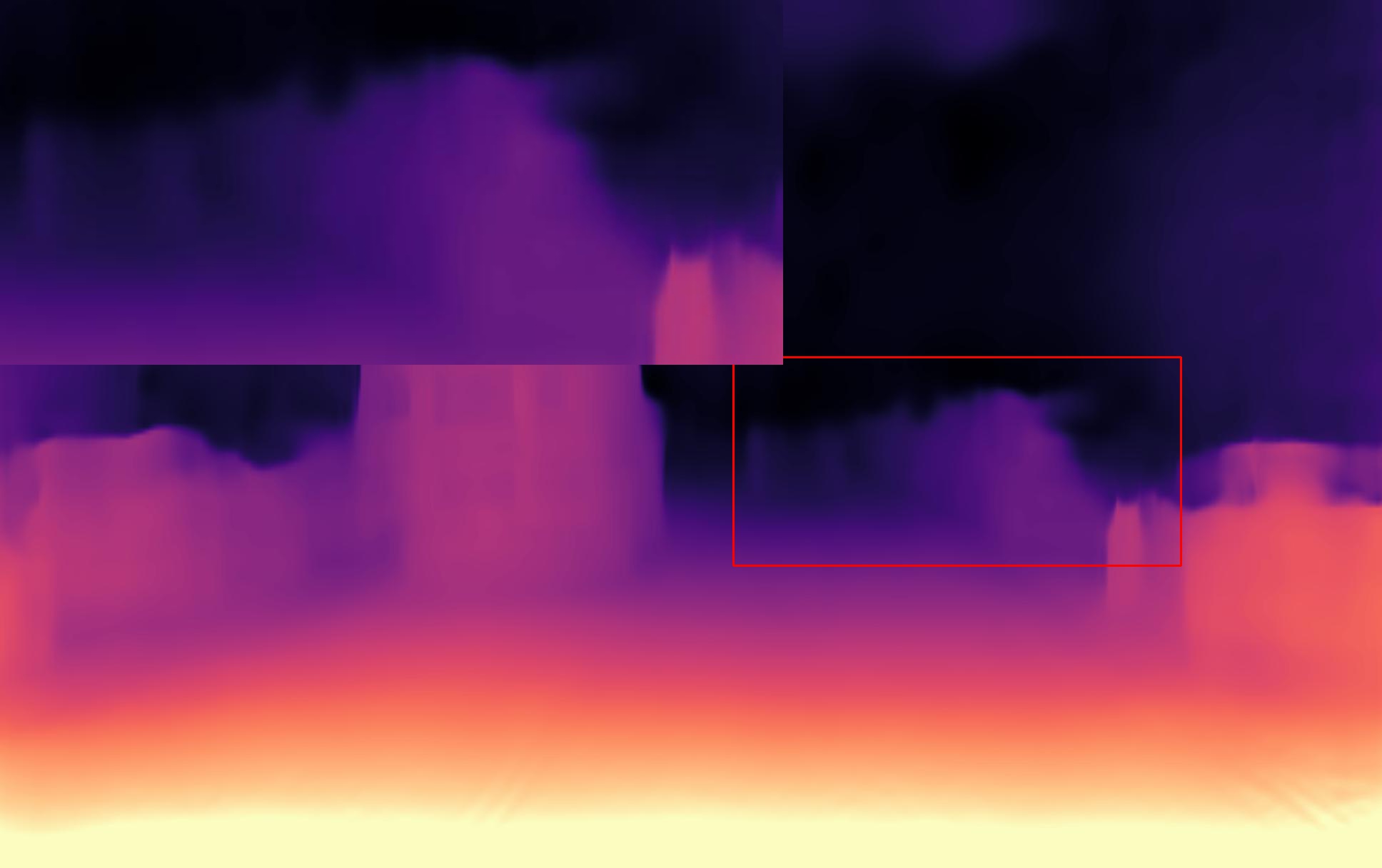} & \includegraphics[width = \textwidth, height = 0.62\textwidth]{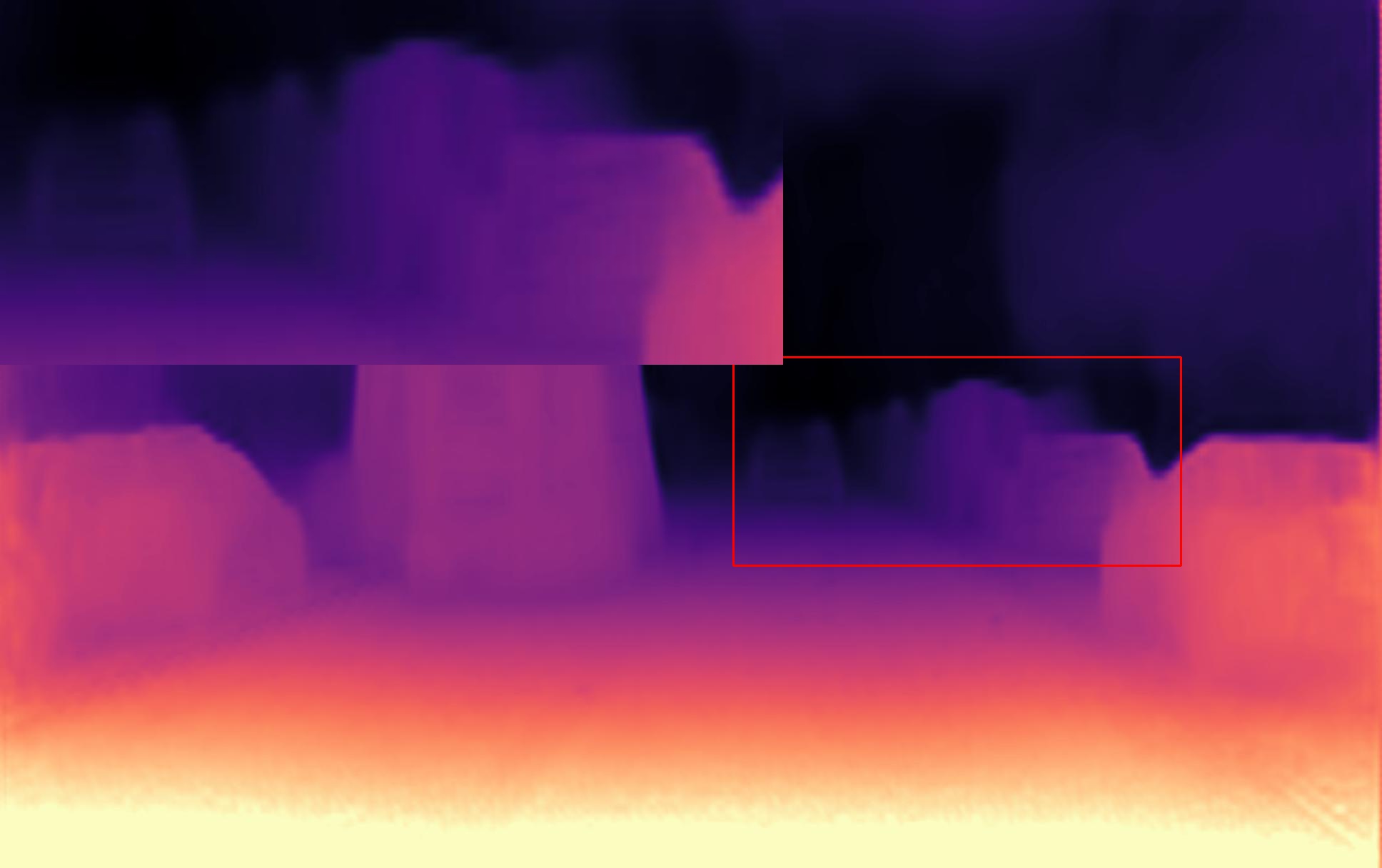} \\ 
\includegraphics[width = \textwidth, height = 0.75\textwidth]{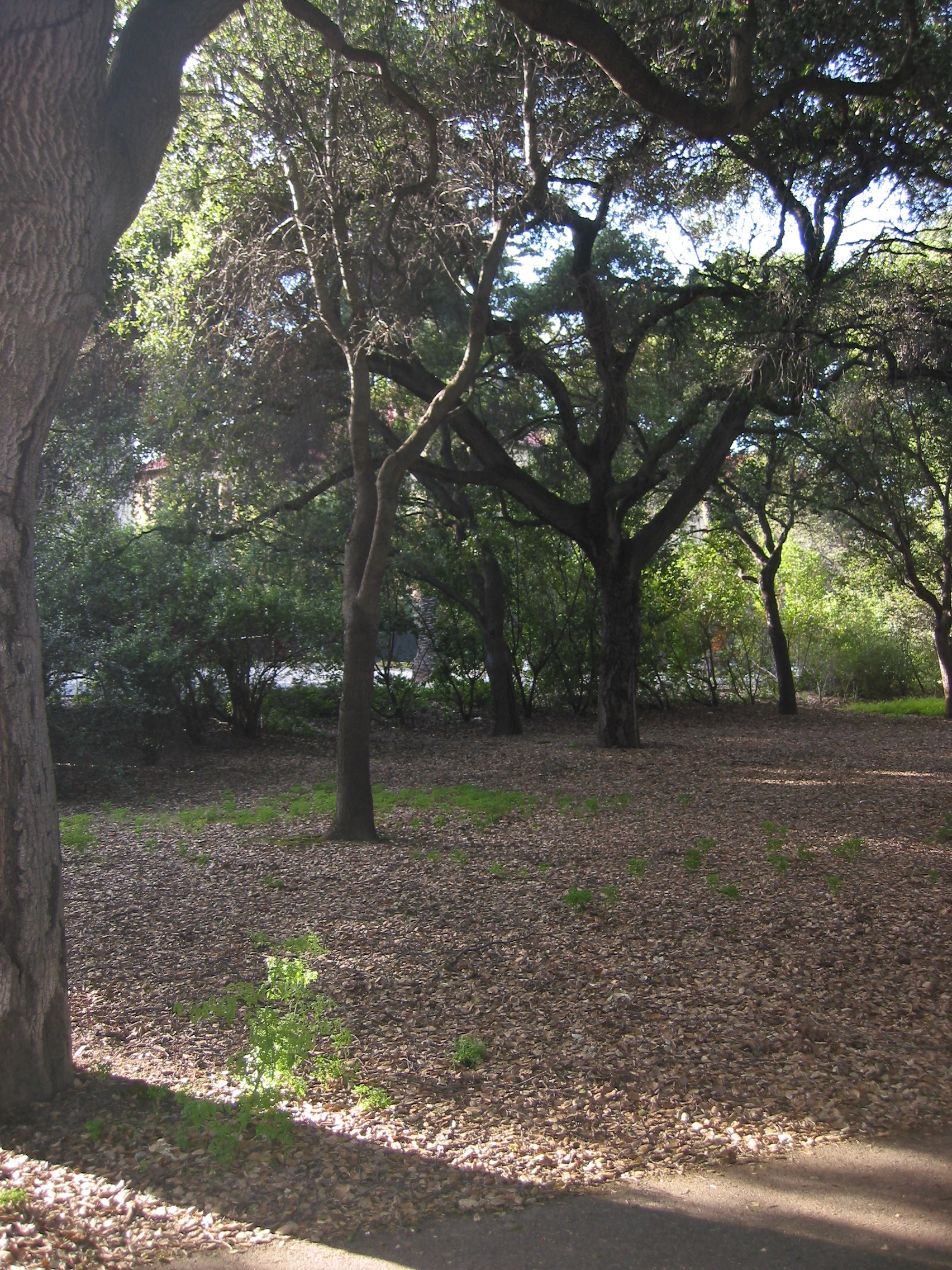} & \includegraphics[width = \textwidth, height = 0.75\textwidth]{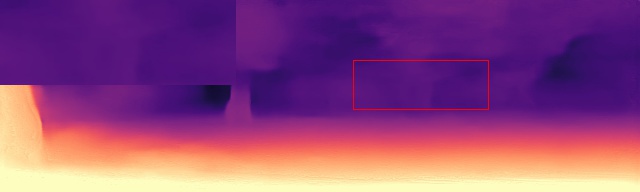} & \includegraphics[width = \textwidth, height = 0.75\textwidth]{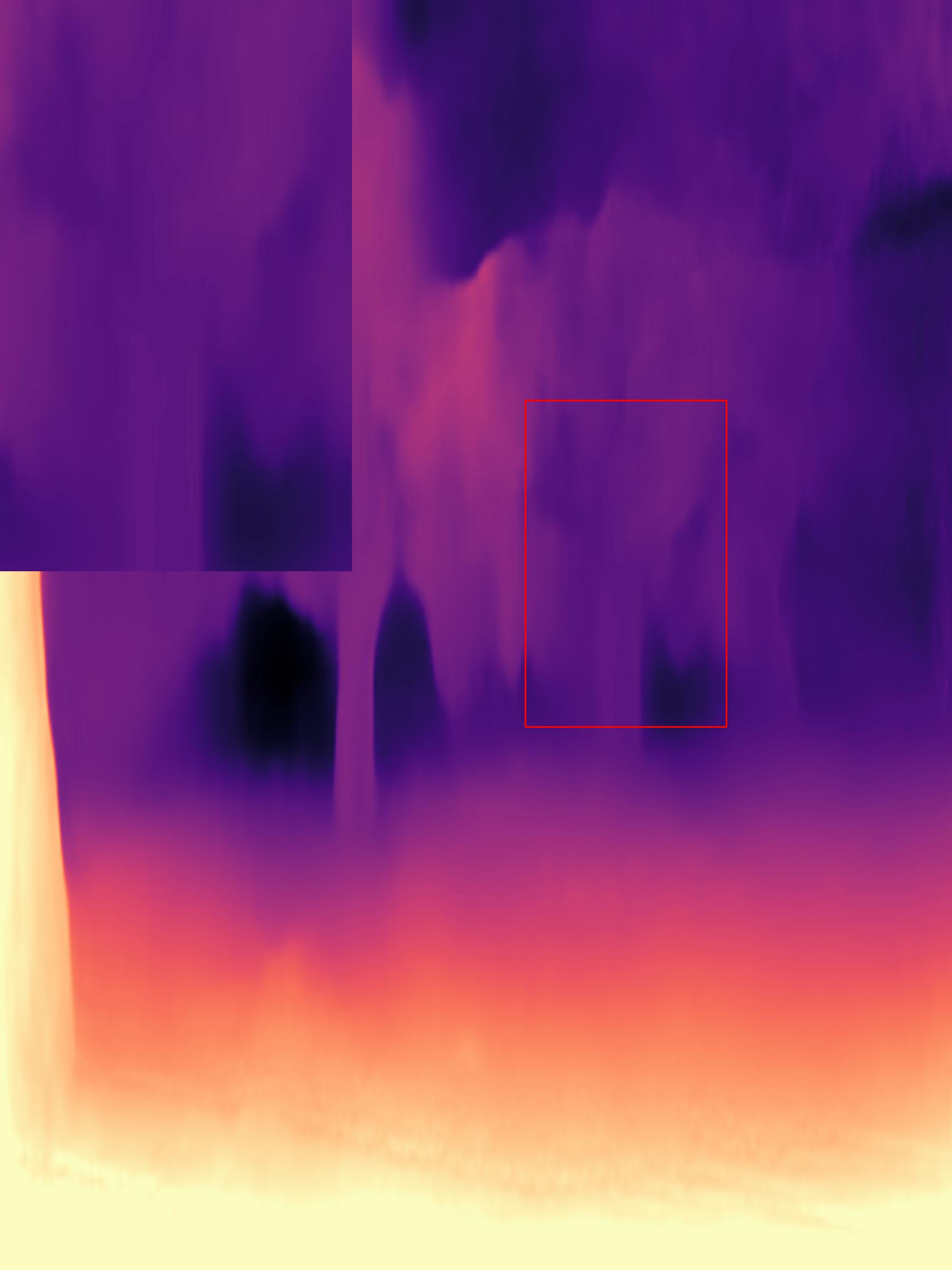} & \includegraphics[width = \textwidth, height = 0.75\textwidth]{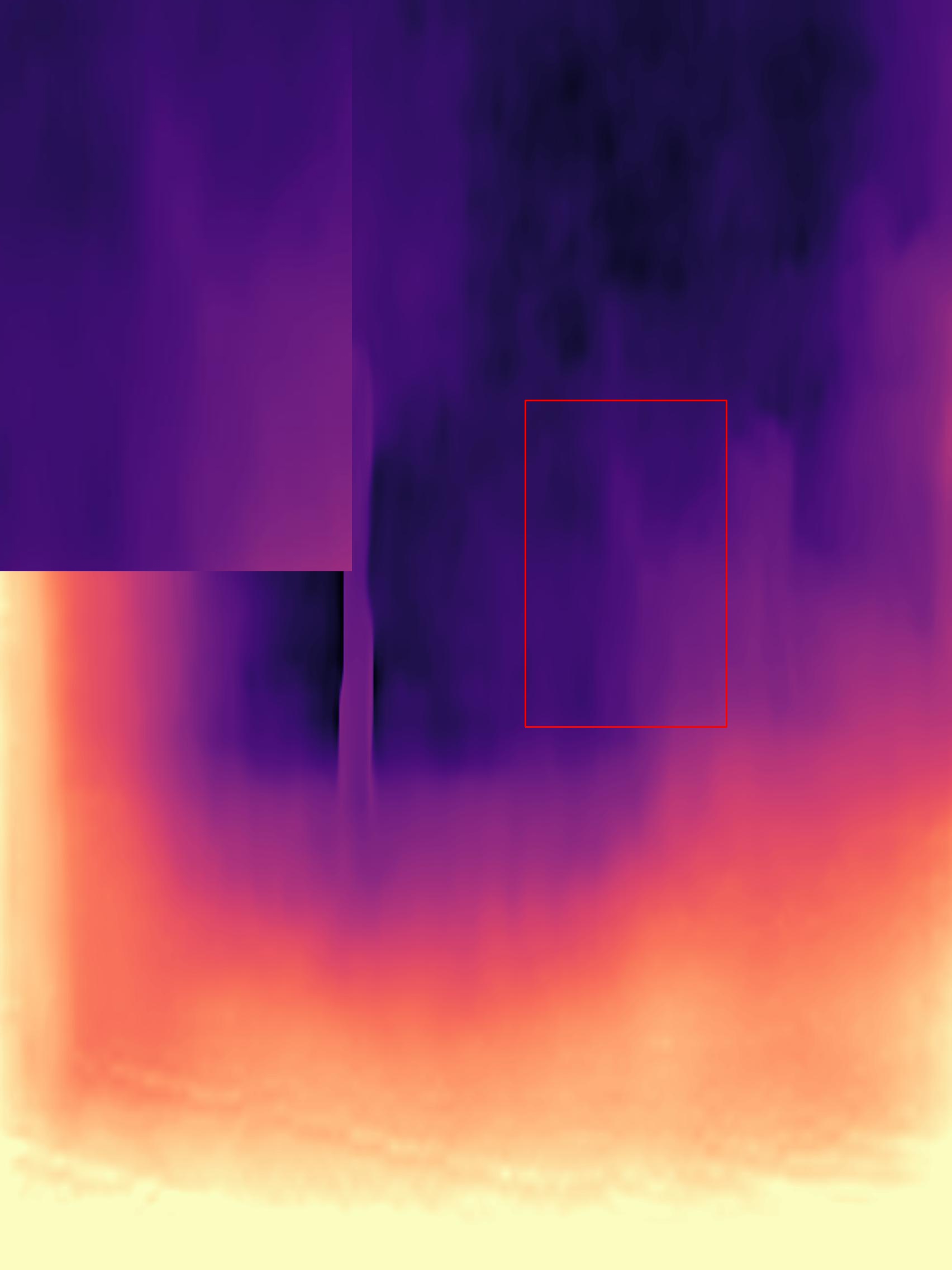} & \includegraphics[width = \textwidth, height = 0.75\textwidth]{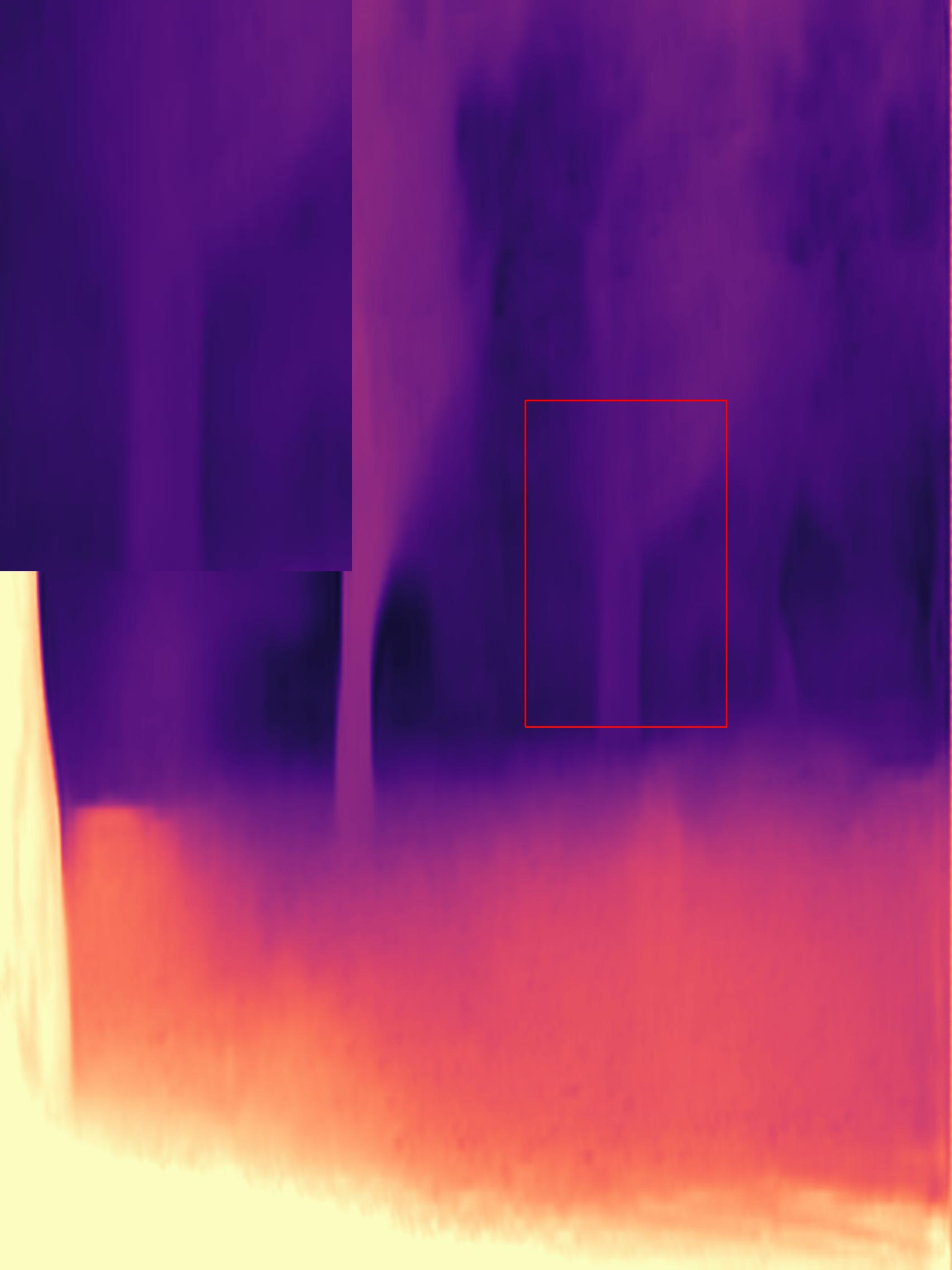} \\ 
\includegraphics[width = \textwidth, height = 0.87\textwidth]{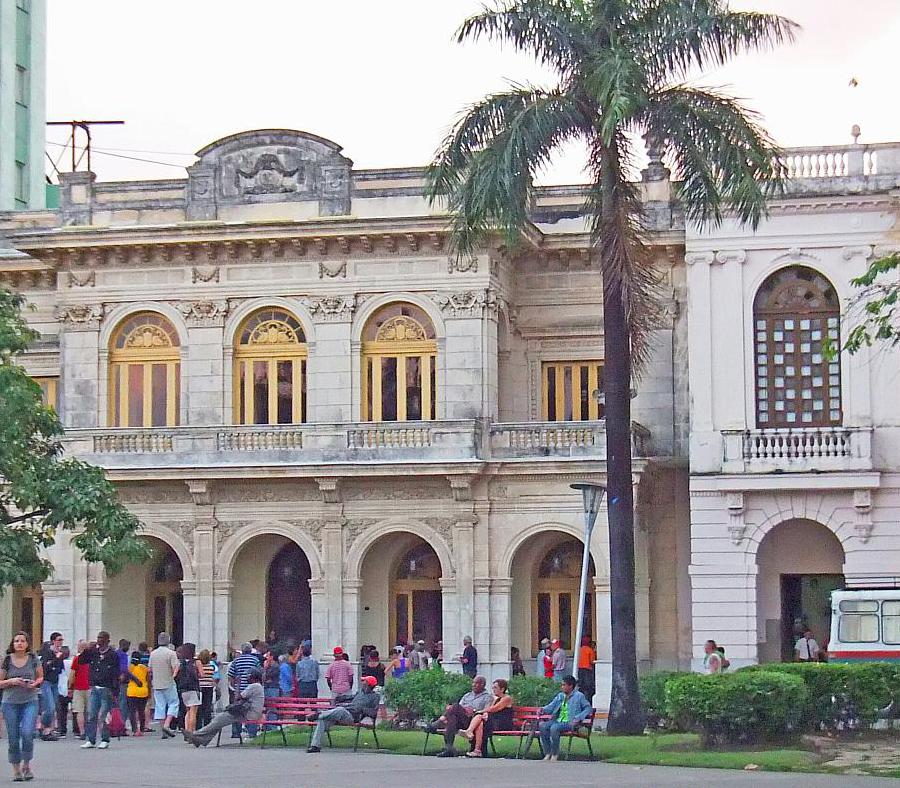} & \includegraphics[width = \textwidth, height = 0.87\textwidth]{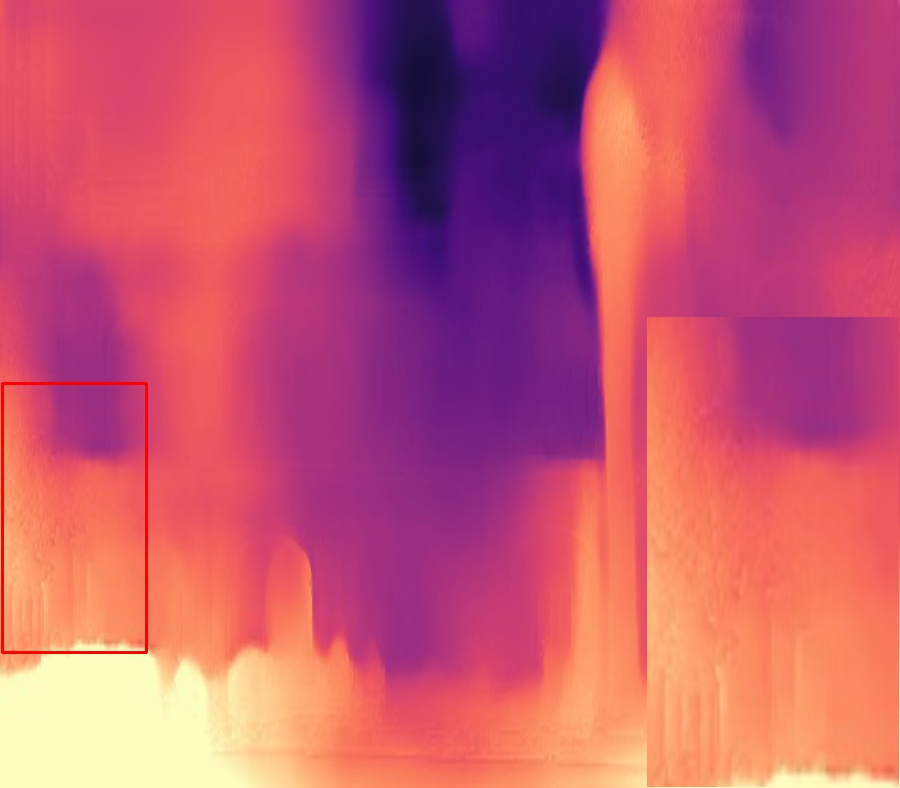} & \includegraphics[width = \textwidth, height = 0.87\textwidth]{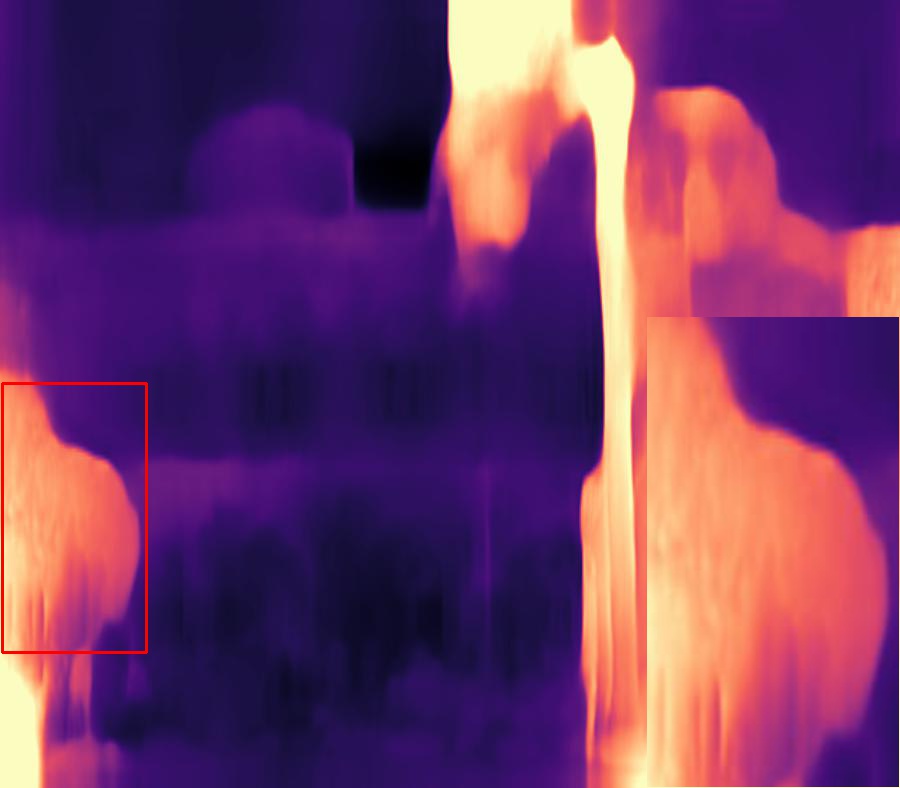} & \includegraphics[width = \textwidth, height = 0.87\textwidth]{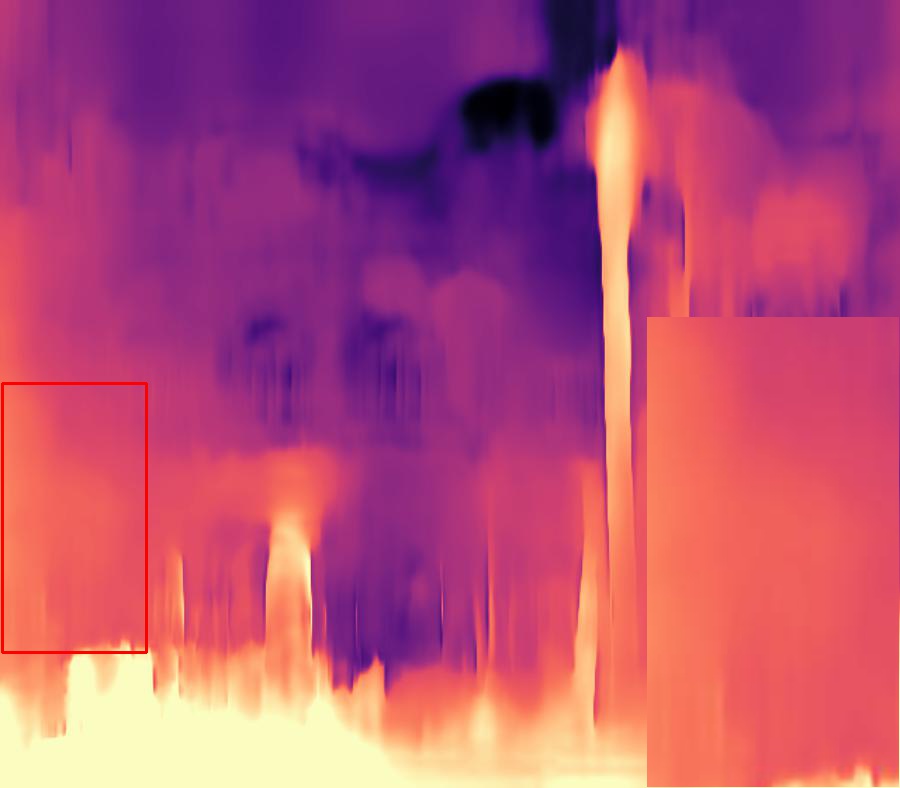} & \includegraphics[width = \textwidth, height = 0.87\textwidth]{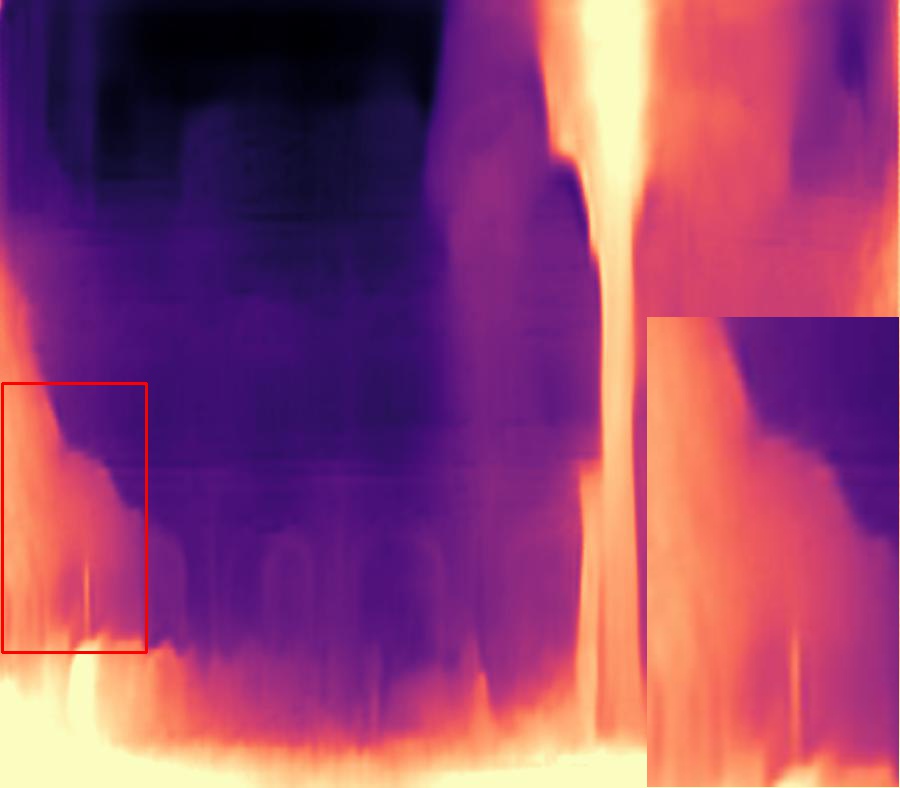} \\
 
\Huge{Input} & \Huge{Packnet-SfM} \cite{Guizilini_2020_CVPR} & \Huge{Monodepth2} \cite{monodepth2} & \Huge{Jonston \textit{et al.}} \cite{Johnston_2020_CVPR} & \Huge{\textbf{Ours}}
\end{tabular}%
}

\caption{Qualitative comparisons on selected test images from the DDAD \cite{Guizilini_2020_CVPR} (Rows 1,2), Make3D \cite{NIPS2005_17d8da81} \cite{4531745} (Row 3), HR-WSI \cite{Xian_2020_CVPR} (Row 4) datasets.}
\label{fig:qual2}
\vspace{-2mm}
\end{figure*}

We have visualized the depth map predicted by our approach and SOTA Packnet-SfM \cite{Guizilini_2020_CVPR}, Johnston \textit{et al.} \cite{Johnston_2020_CVPR}, Monodepth2 \cite{monodepth2} in Fig. \ref{fig:qual}. e have used the models trained on KITTI dataset  \cite{6248074} and directly tested on DDAD \cite{Guizilini_2020_CVPR}, Make3D \cite{NIPS2005_17d8da81} \cite{4531745}, HR-WSI \cite{Xian_2020_CVPR} datasets (Fig. \ref{fig:qual2}). As highlighted in the figures, existing works struggle to accurately predict the depth maps for complex regions or objects and their boundaries. For example, for the 4th column, Fig. \ref{fig:qual}, existing works often fail to distinguish between the vehicle boundary and the background. Similarly, for the 4th row, Fig. \ref{fig:qual2}, they struggle to predict the depth of the highlighted tree region properly. In comparison, our approach is able to gather useful global information from the other regions of the image (e.g., other green tree regions), which helps in accurately identifying an object appearing in front of a complex background. For the 1st row, Fig. \ref{fig:qual2}, our approach produces the local boundaries of the highlighted road sign more accurately, demonstrating its superior ability in processing local scene information, as well.

\subsection{Ablation Studies}
We perform the following experiments, as reported in Table \ref{tab:ablation} on the KITTI dataset \cite{6248074}. Net1: Backbone vision transformer encoder-decoder architecture. Net2: Net1 + low-resolution local encoder branch. Net3: Net2 + high-resolution local encoder branch. Net4: We replace the atrous convolution based fusion block in Net3 with simple addition of features from the three encoder branches. Net5: Net3 + spatial masking in the atrous-fusion module.
\newline \begin{minipage}[]{0.6\textwidth}
\centering
\captionsetup{font=scriptsize}
\captionof{table}{Network Analysis on \cite{6248074}.}
\label{tab:ablation}
\resizebox{.99\textwidth}{!}{%
    \begin{tabular}{|c||c|c|c|c|c|c||c|c|c|c|}   
        \hline
        \textbf{} & \textbf{LRL}  & \textbf{TG} &  \textbf{HRL} & \textbf{Sum} & \textbf{Atr.} & \textbf{Mask} & \textbf{Abs Rel} & \textbf{Sq Rel} & \textbf{RMSE} & \textbf{R. log}  \\
        \hline
        Net1 & & \checkmark & & & & & 0.123 & 0.937 & 4.917 & 0.201  \\ 
        Net2 & \checkmark & \checkmark & & & \checkmark & & 0.113 & 0.791 & 4.637 & 0.188\\ 
        Net3 & \checkmark & \checkmark & \checkmark &  & \checkmark  & & 0.11 & 0.754 & 4.602 & 0.184 \\ 
        Net4 & \checkmark & \checkmark & \checkmark &  \checkmark & & & 0.113 & 0.807 & 4.641 & 0.188 \\ 
        \textbf{Net5} & \checkmark & \checkmark & \checkmark &  & \checkmark & \checkmark & 0.112 & 0.75 & 4.528 & 0.187\\ 
        \hline 
    \end{tabular}%
    }%
    
\end{minipage}
\begin{minipage}[]{0.4\textwidth}
\centering
\captionsetup{font=scriptsize}
    \captionof{table}{Efficiency analysis on \cite{6248074}.}
    \label{tab:eff}
\resizebox{.99\textwidth}{!}{%
    \begin{tabular}{|c||c|c||c|c|c|c|}   
        \hline
        \textbf{} & \textbf{Time}  & \textbf{Params} &  \textbf{Abs Rel} & \textbf{Sq Rel} & \textbf{RMSE} & \textbf{R. log}  \\
        \hline
        \cite{Johnston_2020_CVPR} & \textbf{0.04s} & \underline{38.3M} & \textbf{0.106} & 0.861 & 4.699 & 0.185   \\ 
        \cite{Guizilini_2020_CVPR} & 1.91s & 129M & \underline{0.107} & \underline{0.802} & \underline{4.538} & 0.186\\ 
        \hline
        CNN\_L & 0.20s & 91M & 0.115 & 0.826 & 4.658 & 0.188 \\ 
        TxF\_L & \underline{0.17s} & 82.5M & 0.12 & 0.858 & 4.843 & 0.198\\ 
        \hline
        Ours\_S & 0.21s & \textbf{32.7M} & 0.112 & 0.81 & 4.561 & \textbf{0.184}\\
        Ours\_L & 0.23s & 67.4M & 0.112 & \textbf{0.75} & \textbf{4.528} & \underline{0.187}\\
        \hline 
    \end{tabular}%
    }%

\end{minipage}
The strong performance of our baseline (Net1) shows the utility of transformer-based networks for the current task. For Net2, we introduce the low-resolution local encoder branch in parallel, which processes the image using consecutive convolutional and downsampling operations. We fuse the output of the two encoder branches using the atrous-convolution-based fusion module without any masking operation. The inclusion of the LRL branch improves the performance of the baseline network. For Net3, we deploy another parallel encoder branch that maintains the high-resolution and deep feature representation throughout the encoder. The need for such a technique is reflected in the improved quantitative score of Net3. Overall, the significant improvement of Net2 and Net3 over Net1 shows the complementary behavior of the global transformer and local convolutional branches. To analyze the utility of the proposed fusion module, we replace it with a simple addition of encoder features from the three parallel branches in Net4. The inferior performance of Net4 compared to Net3 demonstrates the need for an adaptive fusion module that can extract multi-scale information from different branches and ultimately capture a better representation of the scene. For Net5, our final model, we introduce the spatial masking operation in the fusion module. Every mask element lies between 0 and 1, representing the importance of different feature positions. Elementwise multiplication with these masks allows our network to fuse only the most helpful information from the parallel local branches. The improved score of Net5 over Net3 strengthens our argument.

We have reported the inference time (per image) and the number of parameters of two recent SOTA works (\cite{Johnston_2020_CVPR}, \cite{Guizilini_2020_CVPR}) and different ablations of our network in Table \ref{tab:eff}. Although our network is slightly heavier than \cite{Johnston_2020_CVPR} due to the three parallel branches, it is much lighter and faster than \cite{Guizilini_2020_CVPR}, whose accuracy is closest to our work. To analyze the role of network complexity for the current task, we train two large baseline networks: TxF\_L and CNN\_L, with a higher number of parameters than our final model (Ours\_L). TxF\_L is a pure transformer network based on \cite{zamir2022restormer}, and CNN\_L is a pure CNN network based on \cite{monodepth2}. These baselines' performance is much inferior to our model, despite having similar or more parameters and runtime. Next, we train a lightweight version of our network (Ours\_S) that achieves comparable performance to our final model (Ours\_L) while having the least number of parameters among all the baselines and SOTA works. These experiments demonstrate that the performance improvement is primarily due to the adaptive fusion of multi-scale information from the three branches, which is the key contribution of our work.

\begin{center}
  \begin{figure}[h]
  \centering
  \vspace{-3mm}
    \resizebox{11.5cm}{!}{    
      \begin{tabular}{c c c}         
        \includegraphics[trim={2.1cm 4.15cm 1.7cm 4.35cm},clip,width = 0.3\textwidth]{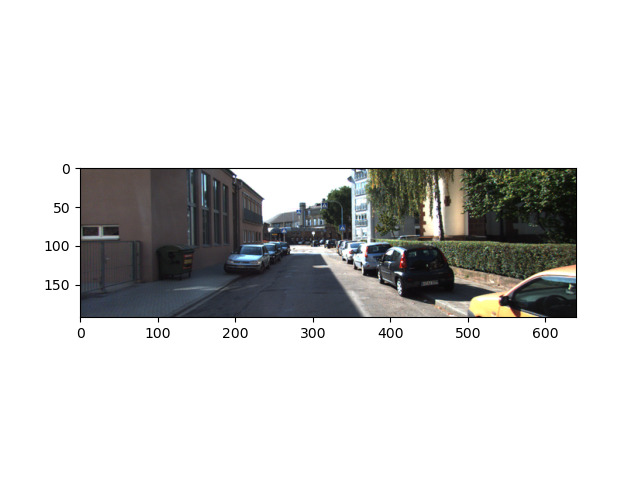}& 
        \includegraphics[trim={2.1cm 4.15cm 1.7cm 4.35cm},clip,width = 0.3\textwidth]{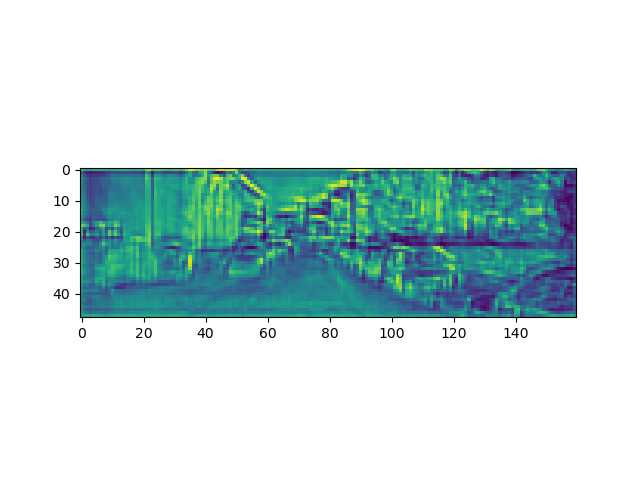}& 
        \includegraphics[trim={2.1cm 4.15cm 1.7cm 4.35cm},clip,width = 0.3\textwidth]{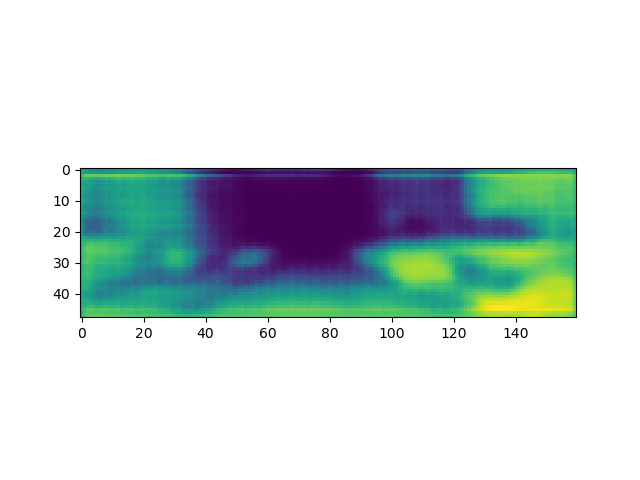} \\
        \includegraphics[trim={2.1cm 4.15cm 1.7cm 4.35cm},clip,width = 0.3\textwidth]{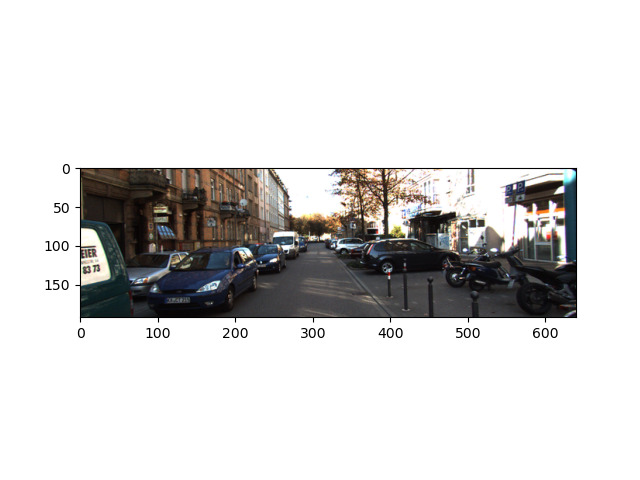}& 
        \includegraphics[trim={2.1cm 4.15cm 1.7cm 4.35cm},clip,width = 0.3\textwidth]{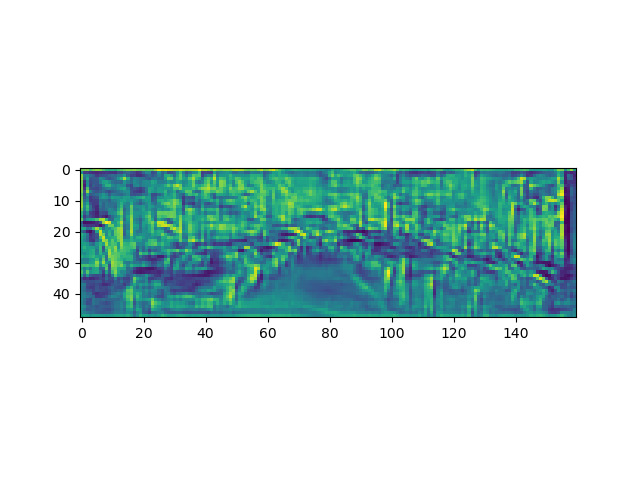}& 
        \includegraphics[trim={2.1cm 4.15cm 1.7cm 4.35cm},clip,width = 0.3\textwidth]{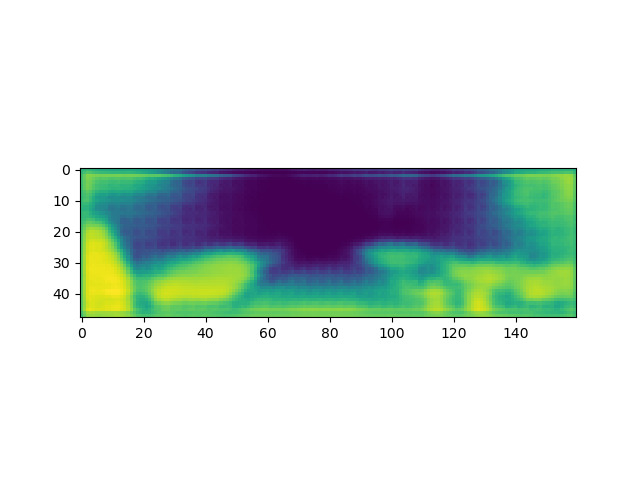} \\
        \includegraphics[trim={2.1cm 4.15cm 1.7cm 4.35cm},clip,width = 0.3\textwidth]{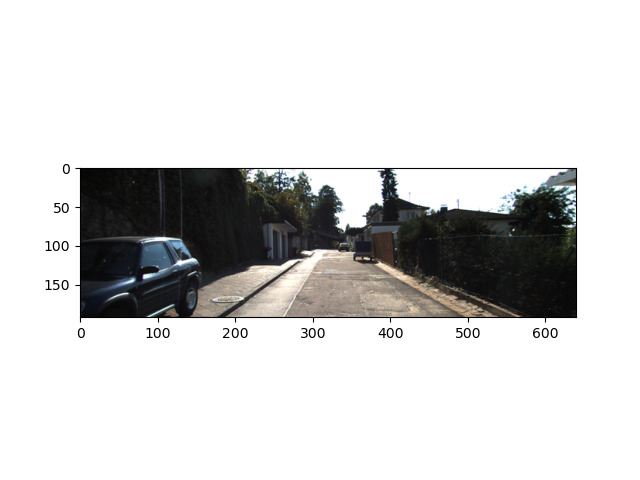}& 
        \includegraphics[trim={2.1cm 4.15cm 1.7cm 4.35cm},clip,width = 0.3\textwidth]{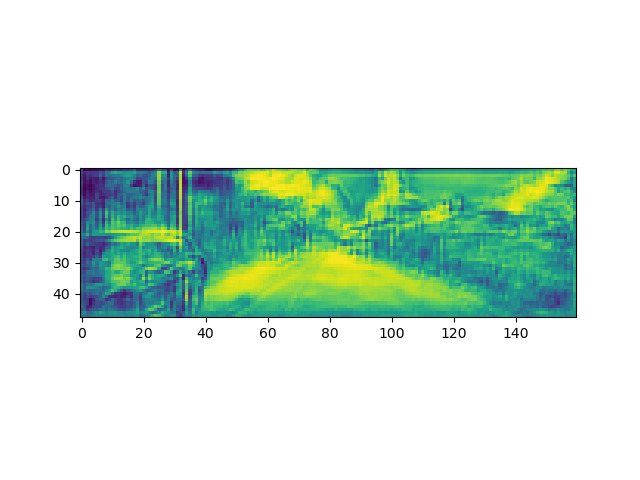}& 
        \includegraphics[trim={2.1cm 4.15cm 1.7cm 4.35cm},clip,width = 0.3\textwidth]{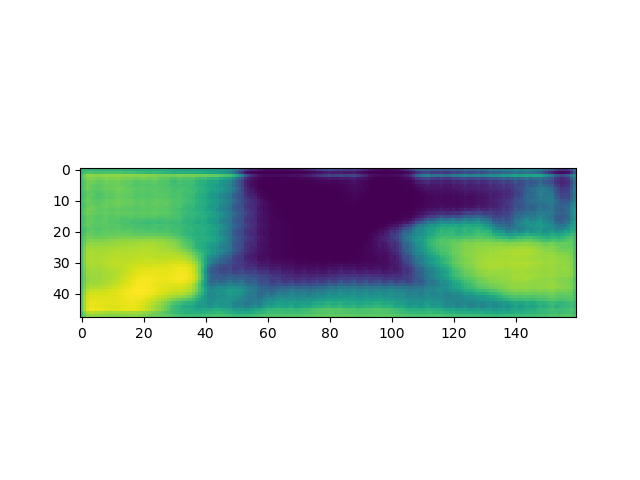} \\
         Input Image & $M^{LRL}$ & $M^{HRL}$\\
       \end{tabular}}
    \caption{Visualization of the spatial mask ($M$) for different encoder branches.}
    \label{fig:mask}
    \vspace{-6mm}
    \end{figure}
\end{center}

We further visualize the spatial mask of Net4 in Fig. \ref{fig:mask} for an intuitive understanding. The masks for the low-resolution and high-resolution local branches (i.e., $M^{LRL}$ and $M^{HRL}$) are shown in Columns 2 and 3, respectively. The spatial masks visually correlate with crucial spatial locations, for example, different objects and structures. Also, the difference between $M^{LRL}$ and $M^{HRL}$ demonstrates the distinct representations learned by the two branches. Interestingly,  $M^{LRL}$ is uniformly distributed throughout the scene while highlighting the relevant parts. This correlates with our intuition that LRL primarily gathers high-level information from the entire frame. In contrast, $M^{HRL}$ is more focused on foreground objects where the depth variation is significant. We argue that HRL excels in handling challenging regions with high-depth variations. Such adaptive ability can be considered crucial to the observed performance improvement of our hybrid network.  
\section{Conclusion}
We have presented a hybrid transformer-based framework for self-supervised monocular depth estimation, achieving improvements over prior arts. Most existing works use a convolutional architecture that fails to satisfactorily model the long-range dependencies between different regions of an image. Due to successive downsampling operations, such designs often fail to preserve the detailed pixel information. On the other hand, using a pure transformer architecture might lack fine-grained local depth information and boundaries. To incorporate the strength of CNNs with the transformer model, we introduce a hybrid framework that extracts helpful information from three encoder branches operating at different spatial resolutions. Complimentary information from the branches is fused adaptively using spatial masks. It would be interesting to extend such a hybrid framework to other image-to-image tasks, such as image enhancement, which would be addressed in future works.
%
%
\bibliographystyle{splncs04}
\bibliography{egbib}
\end{document}